%% file: main.tex
\documentclass[runningheads]{llncs}

\usepackage{eccv}

\usepackage{eccvabbrv}

\usepackage{graphicx}
\usepackage{booktabs}
\usepackage{multirow}%
\usepackage{amsmath,amssymb,amsfonts}%
\usepackage{mathrsfs}%
\usepackage[title]{appendix}%
\usepackage{xcolor}%
\usepackage{textcomp}%
\usepackage{manyfoot}%
\usepackage{algorithm}%
\usepackage{algorithmicx}%
\usepackage{algpseudocode}%
\usepackage{listings}%
\usepackage{tabularx}
\usepackage{subcaption}
\usepackage{graphicx}
\usepackage{bbm}
\usepackage{wrapfig,lipsum,booktabs}
\usepackage{multirow}
\usepackage{makecell}
\usepackage{multicol}
\usepackage{colortbl}
\usepackage{bbding}
\usepackage{amssymb}
\usepackage{pifont}

\definecolor{mygray}{gray}{.9}%
\newcommand{\cmark}{\ding{51}}%
\newcommand{\xmark}{\ding{55}}%

\usepackage[accsupp]{axessibility}  

\usepackage{hyperref}

\usepackage{orcidlink}

\begin{document}

\title{COCA: Classifier-Oriented Calibration via Textual Prototype for Source-Free Universal Domain Adaptation} 

\titlerunning{COCA: Classifier-Oriented Calibration via Textual Prototype for SF-UniDA}

\author{Xinghong Liu\inst{1} \and
Yi Zhou\inst{1}\thanks{Corresponding author} \and
Tao Zhou\inst{2} \and
Chun-Mei Feng\inst{3} \and
Ling Shao\inst{4}}

\authorrunning{X.~Liu et al.}

\institute{School of Computer Science and Engineering, Southeast University, Nanjing, China \\ \email{\{xhoml158,yizhou.szcn\}@gmail.com}\and
School of Computer Science and Engineering, Nanjing University of Science and Technology, Nanjing, China
\\ \email{taozhou.ai@gmail.com} \and
Institute of High Performance Computing (IHPC), Agency for Science, Technology and Research (A*STAR), Singapore
\\ \email{strawberry.feng0304@gmail.com} \and
UCAS-Terminus AI Lab, University of Chinese Academy of Sciences, Beijing, China \\ \email{ling.shao@ieee.org}}
\maketitle

\begin{abstract}
Universal domain adaptation (UniDA) aims to address domain and category shifts across data sources.
Recently, due to more stringent data restrictions, researchers have introduced source-free UniDA (SF-UniDA).
SF-UniDA methods eliminate the need for direct access to source samples when performing adaptation to the target domain. However, existing SF-UniDA methods still require an extensive quantity of labeled source samples to train a source model, resulting in significant labeling costs.
To tackle this issue, we present a novel plug-and-play \textbf{C}lassifier-\textbf{O}riented \textbf{CA}libration (COCA) method. COCA, which exploits textual prototypes, is designed for the source models based on few-shot learning with vision-language models (VLMs).
It endows the VLM-powered few-shot learners, which are built for closed-set classification, with the unknown-aware ability to distinguish common and unknown classes in the SF-UniDA scenario.
Crucially, COCA is a new paradigm to tackle SF-UniDA challenges based on VLMs, which focuses on classifier instead of image encoder optimization.
Experiments show that COCA outperforms state-of-the-art UniDA and SF-UniDA models.
\end{abstract}

\section{Introduction}
\input{eccv24/1_introduction}

\section{Related Work}
\input{eccv24/2_related_work}

\section{Methodology}
\input{eccv24/3.0_preliminary}
\subsection{Autonomous Calibration via Textual Prototype}
\input{eccv24/3.1_actp}
\subsection{Mutual Information Enhancement by Context Information}
\input{eccv24/3.2_mieci}
\subsection{Model Optimization}
\input{eccv24/3.3_model_optimization}
\subsection{Inference}
\input{eccv24/3.4_inference}

\section{Experiments}
\subsection{Datasets and Evaluation Metric}
\input{eccv24/4.1_datasets_and_evaluation_metric}
\subsection{Comparisons with State-of-the-Art Methods}
\input{eccv24/4.2_comparison_with_sota}
\subsection{Ablation Studies}
\input{eccv24/4.3_further_evaluations}

\section{Conclusion}
\input{eccv24/5_conclusion}

\newpage
\appendix

In this appendix, we provide more details of our approach, such as additional experiment results, extensive implementation details, and discussion.

This supplementary is organized as follows:
\renewcommand{\labelitemii}{$\circ$}

\begin{itemize}
\setlength{\itemindent}{-0mm}
    \item Additional Experiment Results
    \item Implementation Details
    \begin{itemize}
        \setlength{\itemindent}{-0mm}
        \item Source Model Details
        \item Silhouette Score
        \item Pseudo Code
        \item Baseline Details
    \end{itemize}
    \item Discussion
    \begin{itemize}
        \setlength{\itemindent}{-0mm}
        \item K-means Clustering Invocations
        \item Potential Societal Impact
    \end{itemize}
\end{itemize}{}

\section{Additional Experiment Results}
\input{eccv24/appendix/experiment_results}

\section{Implementation Details}
\input{eccv24/appendix/implementation_details}

\section{Discussion}
\input{eccv24/appendix/discussion}
%
%
\bibliographystyle{splncs04}
\bibliography{eccv_main}
\end{document}

%% file: eccv24/1_introduction.tex
Training deep neural networks (DNNs) on custom datasets can achieve outstanding performance when the models are employed on $i.i.d.$ datasets.
However, the trained DNNs are likely to underperform if the \textcolor{black}{test} dataset (target domain) exhibits a large domain shift compared to the \textcolor{black}{training} dataset (source domain).
To address the performance degradation in unlabeled target domains caused by domain shift, researchers have \textcolor{black}{studied} unsupervised domain adaptation (UDA) \cite{DANN,DDAN,Zheng2021_IJCV,MIC, Roy_2019_CVPR} methods.
Vanilla UDA methods are designed for the closed-set scenario.
In real-world situations, category shift may arise, \textcolor{black}{including} open-set domain adaptation (OSDA) \cite{OSBP,STA, yizhou2023,UADAL,SPL-OSDA}, partial domain adaptation (PDA) \cite{AR, BA3US, PADA}, and open-partial domain adaptation (OPDA) \cite{UniOT,SSM}.
Thus, universal domain adaptation (UniDA) \cite{DANCE,DCC,OVANet,SPA} \textcolor{black}{is} introduced to address such uncertain domain and category shifts, meaning it must handle \textcolor{black}{arbitrary} situations that may arise in OSDA, PDA, or OPDA.
More recently, due to stricter data restrictions, source-free universal domain adaptation (SF-UniDA) \cite{GLC} has been proposed. When performing SF-UniDA, it assumes that only the trained source model is available, rather than providing source samples. 
However, the existing SF-UniDA method \cite{GLC} still requires training the source model using numerous labeled source samples, leading to substantial labeling costs.

\begin{figure}[t]
\centering
     \includegraphics[width=0.85\textwidth]{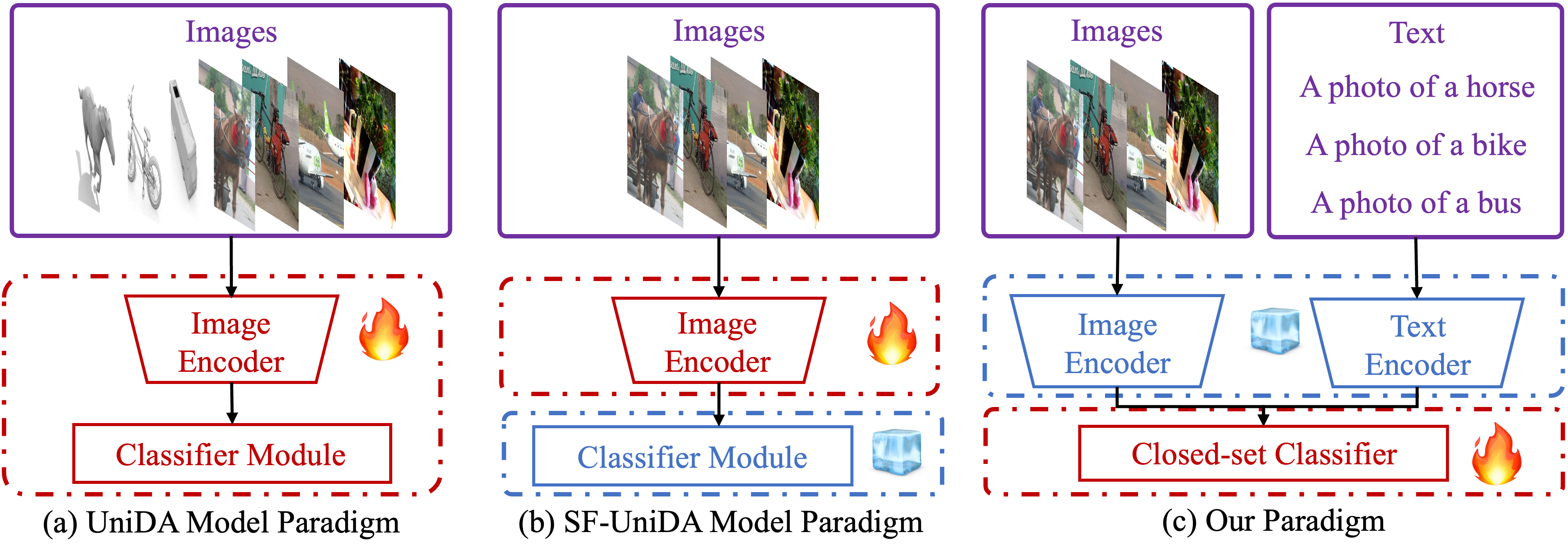}
     \caption{(a) UniDA methods adapt both the image encoder and classifier module. (b) SF-UniDA methods freeze the classifier module and adapt the image encoder. (c) We leverage and freeze image and text encoders and adapt the closed-set classifier.}
     \label{Fig:Pipeline_comparison}
\end{figure}

Recently, the emergence of vision-language models (VLMs) such as CLIP \cite{CLIP}, inspires us that labeling costs of source samples can be minimized by utilizing \textcolor{black}{the strong representation ability of these models}. The VLMs have become the new foundational engine for few-shot learning \cite{WISE-FT,CoOp,gao2021clipadapter,cross-modal_adaptation}.
However, existing \textcolor{black}{VLM-powered few-shot learners} are primarily designed for closed-set scenarios and $i.i.d.$ domain distribution. They experience substantial performance degradation when confronted with domain and category shifts. In this paper, we propose a plug-and-play method to calibrate the few-shot learning models, such as linear probe CLIP \cite{CLIP}, CLIP Adapter \cite{gao2021clipadapter}, and cross-modal linear probing \cite{cross-modal_adaptation}, \textcolor{black}{to tackle the SF-UniDA challenge. Furthermore, we investigate the zero-shot learning problem within the context of SF-UniDA tasks. Specifically, we adapt zero-shot classifiers such as single linear layer \cite{CLIP} or the adapter module \cite{gao2021clipadapter} to target domains.} Traditional approaches in UniDA and SF-UniDA have predominantly concentrated on transferring specific knowledge from the source to the target domain. \textcolor{black}{This is because the traditional image encoders such as ResNet \cite{ResNet}, which are pre-trained on the ImageNet \cite{ImageNet} dataset, do not inherently encapsulate knowledge of both the source and target domains.} We argue that the image and text encoders within the VLMs, having undergone pre-training on extensive image-text pair datasets, intrinsically encapsulate knowledge pertinent to both domains. \textcolor{black}{As depicted in \cref{Fig:Pipeline_comparison}, in contrast to conventional UniDA or SF-UniDA methods that primarily concentrate on image encoder optimization to transfer source domain knowledge, we present a new paradigm focusing on classifier optimization to tackle the SF-UniDA challenge.}

\begin{figure}[t]
\centering
     \begin{subfigure}[b]{0.32\textwidth}
         \centering
         \includegraphics[width=\textwidth]{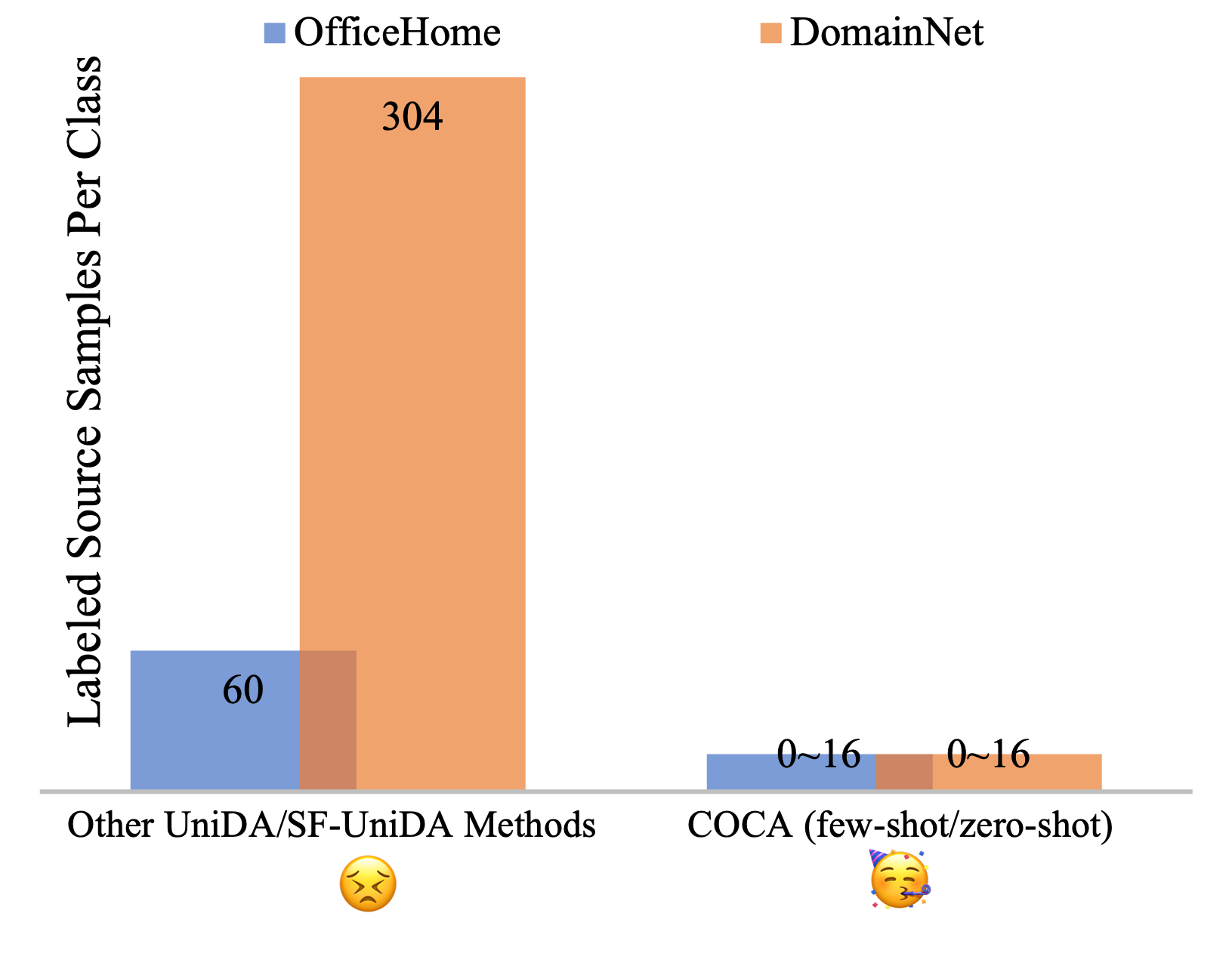}
         \caption{}
         \label{Fig:Comparison_with_UniDA}
     \end{subfigure}
     \hfill
     \begin{subfigure}[b]{0.32\textwidth}
         \centering
         \includegraphics[width=\textwidth]{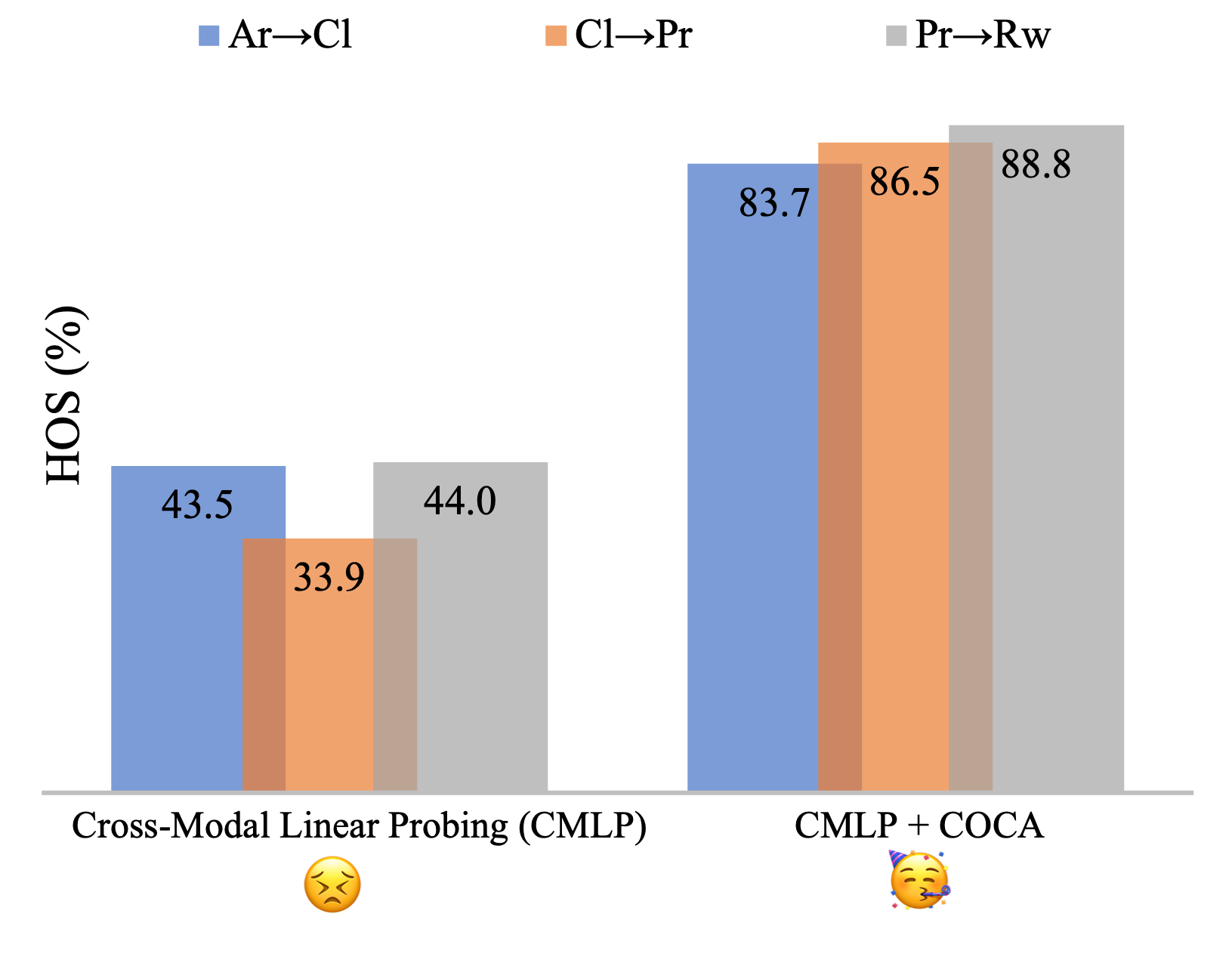}
         \caption{}
         \label{Fig:Comparison_with_Cross-Modal}
     \end{subfigure}
     \hfill
     \begin{subfigure}[b]{0.32\textwidth}
         \centering
         \includegraphics[width=\textwidth]{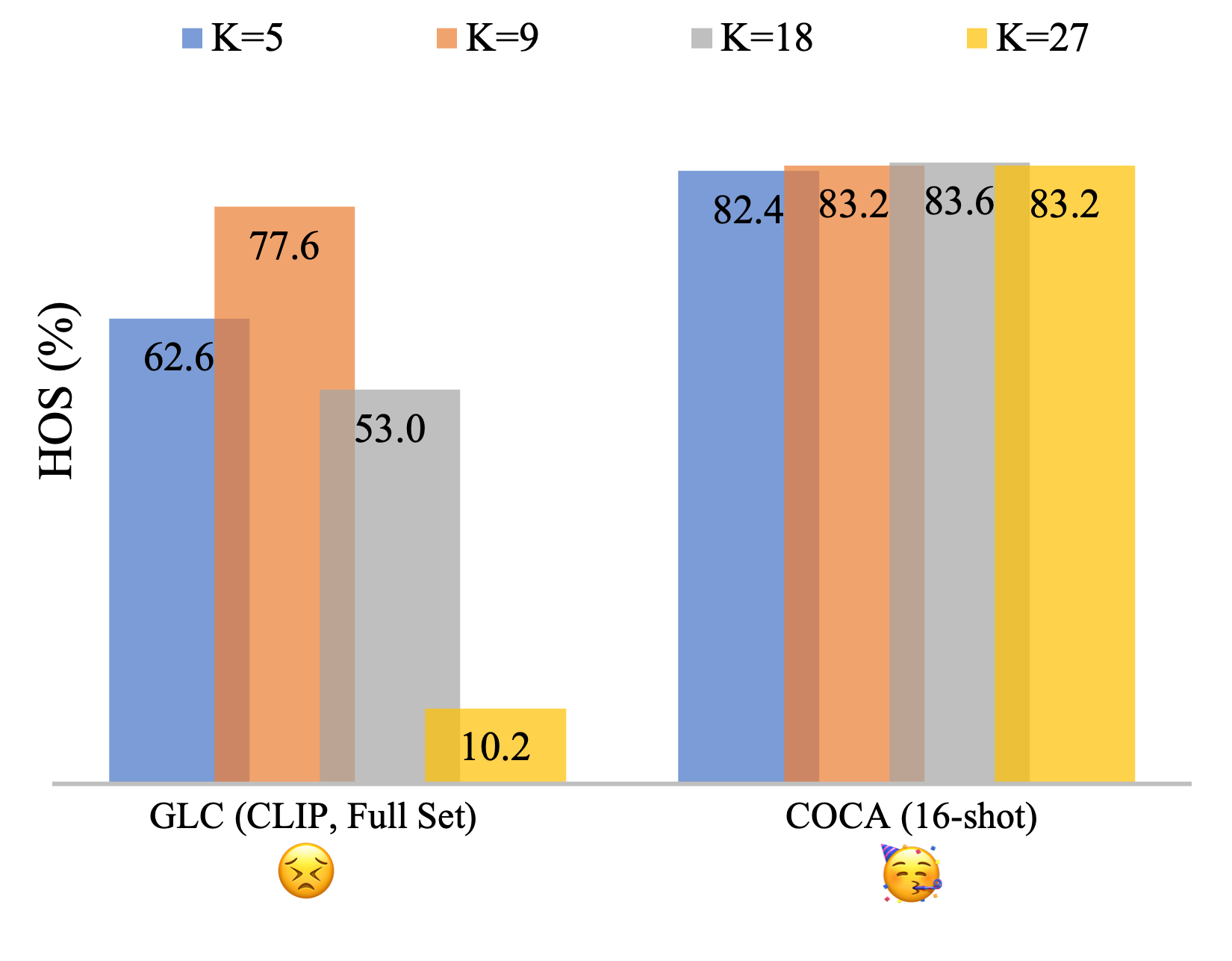}
         \caption{}
         \label{Fig:Comparison_with_GLC}
     \end{subfigure}
     \caption{(a) Our method requires far fewer labeled source samples per class than traditional UniDA/SF-UniDA models. \textcolor{black}{(b) Our plug-and-play method successfully adapts the VLM-powered few-shot learner \cite{cross-modal_adaptation} to new target domains.} (c) COCA exhibits more robustness against variations in the hyperparameter $K$ for K-means and outperforms the earlier SF-UniDA model GLC \cite{GLC}.}
\end{figure}

In order to adapt the closed-set classifier to new target domains, we present a novel plug-and-play classifier-oriented calibration (COCA) method, which \textcolor{black}{exploits} textual prototypes, \textcolor{black}{to endow the VLM-powered few-shot learners with unknown-aware ability.}
To overcome domain and category shifts and enable the source model—specifically, the VLM-powered few-shot learner—to accurately distinguish common and unknown class samples, we introduce a novel autonomous calibration via textual prototype (ACTP) module. 
ACTP utilizes the K-means \cite{K-means} clustering, along with image and text features, to implement \textcolor{black}{self-training to adapt the close-set classifier to new unlabeled target domains.}
Furthermore, to encourage the closed-set classifier to exploit context information in images and enhance the mutual information, we propose the mutual information enhancement by context information (MIECI) module. Our experiments reveal that MIECI favorably influences model performance.

\textcolor{black}{We illustrate our research motivation in \cref{Fig:Comparison_with_UniDA}.} For the SF-UniDA challenge, our approach can be applied to the few-shot/zero-shot learning problems, significantly reducing labeling costs of source samples compared to previous UniDA or SF-UniDA methods. \textcolor{black}{As illustrated in \cref{Fig:Comparison_with_Cross-Modal}, COCA endows VLM-powered few-shot learners with the unknown-aware ability to accurately distinguish common and unknown class samples within target domains.}
Moreover, we have identified two issues: (1) The conventional UniDA and SF-UniDA methods \cite{DCC,GLC}, which utilize prototypes derived from image features—termed image prototypes, unavoidably incorporate domain information into these prototypes. This hinders the mutual information $I(X,V)$ between original images $X$ and prototypes $V$ to be minimized. Consequently, the model performances are suboptimal. (2) The efficacy of the preceding model GLC \cite{GLC}, which employs image prototypes, heavily relies on the empirical selection of appropriate hyperparameter $K$ for K-means clustering. This is because the setting of $K$ critically affects the quality of the image prototypes. To address the two issues, we propose a novel approach wherein text features are employed as prototypes—termed textual prototypes—to minimize the mutual information $I(X,V)$, consequently enhancing the model performance. This approach, grounded in textual prototypes, guarantees more stable performance. Our approach performs better and is insensitive to variations in the hyperparameter $K$ as shown in \cref{Fig:Comparison_with_GLC}.


We conduct experiments on three public benchmarks, each offering sufficient labeled source samples. Previous UniDA and SF-UniDA methods are fully trained using the source sample sets, but we only use few source shots to train the source model in our approach. The results indicate that COCA consistently surpasses state-of-the-art methods over all the benchmarks, even though our source model is trained on few source shots.
To summarize, our contributions are highlighted as follows:
\begin{enumerate}
\item To the best of our knowledge, we are the first to explore few-shot and zero-shot learning problems in the UniDA/SF-UniDA scenario. \textcolor{black}{Specifically, we propose a plug-and-play method for the VLM-powered few-shot learners to adapt them to new target domains and to endow them with unknown-aware ability.  Crucially, we present a new paradigm, which focuses on classifier instead of image encoder optimization, to tackle the SF-UniDA challenge.}
\item To overcome domain and category shifts and enable the closed-set source models to accurately distinguish common and unknown class samples, we present a novel autonomous calibration via textual prototype (ACTP) module. We introduce the mutual information enhancement by context information (MIECI) module to encourage the classifier to exploit context information in image features and enhance the mutual information.
\item Experiments conducted on three public benchmarks under various category-shift settings show the substantial superiority of our approach. Moreover, our study reveals that VLMs have encapsulated knowledge of both the source and target domains, enabling VLM-powered models to autonomously adapt to new target domains.
\end{enumerate}

%% file: eccv24/2_related_work.tex
\textbf{Universal Domain Adaptation.} Li \etal \cite{DCC} defined the universal domain adaptation (UniDA) scenario measuring the robustness of a model to various category shifts. They leveraged domain consensus knowledge to enhance target clustering and discover private categories. OVANet \cite{OVANet} employs a one-vs-all classifier for each source class, determining the "known" or "unknown" class through its output. Conventional UniDA methods require direct access to source samples for domain adaptation. In response to burgeoning data protection policies, the source-free universal domain adaptation (SF-UniDA) is proposed in \cite{GLC}. However, GLC \cite{GLC} still requires an extensive quantity of labeled source samples to develop a source model, resulting in significant labeling costs. Moreover, We observe that the existing UniDA and SF-UniDA paradigms concentrating on image encoder optimization are ill-suitable for VLMs.

\noindent\textbf{Few-Shot Learning.} In the context of few-shot visual classification, a classifier is initially pretrained on a set of base classes \cite{ImageNet} to learn a good feature representation and then finetuned on a limited number of novel class samples. In recent advancements, some few-shot learning methods \cite{gao2021clipadapter,cross-modal_adaptation} based on CLIP \cite{CLIP} have been proposed. However, these methods, specifically designed for $i.i.d.$ datasets and the closed-set scenario, exhibit performance deterioration due to domain and category shifts, making them unsuitable to be directly applied in UniDA/SF-UniDA. Furthermore, it is unfeasible to label few-shot target samples in UniDA/SF-UniDA since the target class set is uncertain. E.g., in OSDA/OPDA scenarios, the quantity of unknown classes remains uncertain, and in scenarios like OPDA/PDA, certain classes might be absent in the target domain.

%% file: eccv24/3.0_preliminary.tex
\begin{figure}[t]
\includegraphics[width=\textwidth]{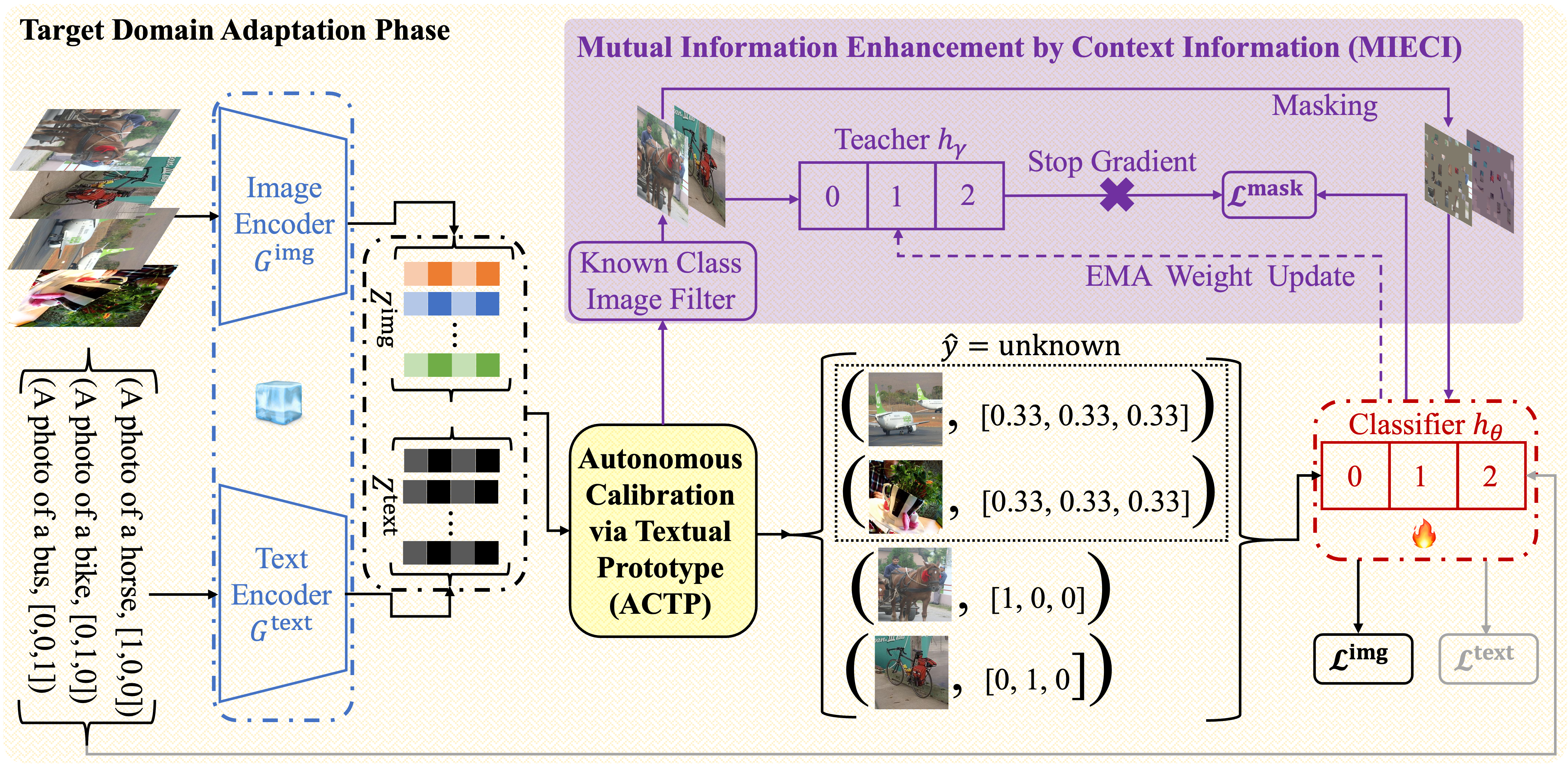}
\caption{Overview of the classifier-oriented calibration (COCA) method. COCA adapts the closed-set classifier $h_\theta$ to the target domain to tackle the SF-UniDA challenge.}
\label{Fig:model_diagram}
\end{figure}

\noindent\textbf{Preliminaries.} We \textcolor{black}{have} a labeled source domain $D^s = \left\{(x^s, y^s):y^s\in C^s\right\}$, where $C^s$ denotes the source class set. 
Each class name in the source domain is associated with its respective ground truth label, denoted as $y_c$. For instance, the class name \verb+horse+ carries the ground truth label $y_c=[1,0,0]$ in \cref{Fig:model_diagram}.
We also have an unlabeled target domain $D^t = \left\{(x^t)\right\}$ with a domain distribution that differs from that of $D^s$.
Assuming $C^t$ is the target class set, the relationship between $C^s$ and $C^t$ in the UniDA scenario can be categorized into $C^s \subset C^t$ (OSDA), $C^t \subset C^s$ (PDA), and $C^s \cap C^t \neq \emptyset, C^s \not \subset C^t, C^t \not \subset C^s$ (OPDA).
The classes in $C^s$ are referred to as known classes, while the classes in $C = C^s \cap C^t$ are termed common classes. The source- and target-private class sets are defined as $\bar{C^s} = C^s \setminus C^t$ and $\bar{C^t} = C^t \setminus C^s$, and the classes in $\bar{C^t}$ are unknown classes.

\noindent\textbf{Overview.} The overview of the classifier-oriented calibration (COCA) approach is depicted in \cref{Fig:model_diagram}. \textcolor{black}{For the few-shot learning problem, we use the VLM-powered few-shot learner trained on the source domain as the source model. In the zero-shot learning problem, the classifier's weights are initialized with the text features.} At the target domain adaptation phase, we adapt the closed-set classifier $h_\theta$ of the source model to the target domain. To enable the classifier to distinguish common and unknown class samples in the target domain, we propose the autonomous calibration via textual prototype (ACTP) module that exploits the close feature distance between text features and image prototypes. We introduce the mutual information enhancement by context information (MIECI) module, which includes a teacher classifier $h_\gamma$, to encourage the classifier $h_\theta$ to exploit context information in images and enhance the mutual information.

%% file: eccv24/3.1_actp.tex
\begin{figure*}[t]
\includegraphics[width=\textwidth]{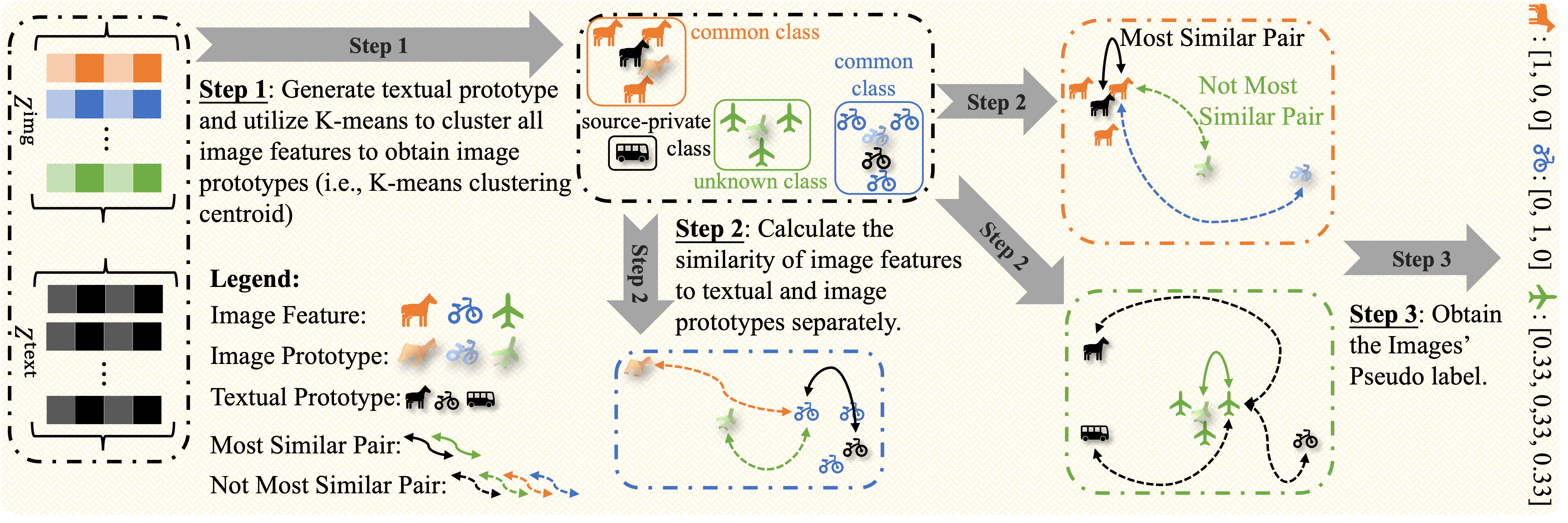}
\caption{\textcolor{black}{Pipeline of the autonomous calibration via textual prototype (ACTP) module. \texttt{horse} and \texttt{bike} are the common classes, \texttt{bus} is the source-private class, and \texttt{airplane} is the unknown class. In Step 3, ACTP generates pseudo labels via \cref{eq:generate_pseudo_label}.}}
\label{Fig:actp_diagram}
\end{figure*}

To enable the closed-set classifier $h_\theta$ to distinguish common and unknown class samples, we introduce the autonomous calibration via textual prototype (ACTP) module. The pipeline of the ACTP module is depicted in \cref{Fig:actp_diagram}.

\textbf{Positive and Negative Prototypes.} \textcolor{black}{Leveraging the close feature distance between text features and image prototypes generated by VLMs, we can identify positive and negative prototypes for known classes using the class names.} To generate positive prototypes for known classes, we input \texttt{a photo of a \{CLS\}} to the text encoder to obtain a text feature set $Z^\text{text}=\{z^\text{text}_c\}^{\left|C^s\right|}_{c=1}$, where $|C^s|$ is the number of source classes and the class tokens \texttt{\{CLS\}} are replaced by specific class names in the source domain, such as \verb+horse+, \verb+bike+, or \verb+bus+. The prediction probability that a target image $x_i$ belongs to a known class $c$ is computed as:
\begin{equation}
p(y=c|x_i)=\frac{\exp(\cos(z^\text{text}_c,z^\text{img}_i)/\mathcal{T})}{\sum^{|C^s|}_{j=1}\exp(\cos(z^\text{text}_j,z^\text{img}_i)/\mathcal{T})},
\label{eq:pred_probability}
\end{equation}
where $\mathcal{T}$ is a temperature parameter, $\cos(\cdot,\cdot)$ denotes cosine similarity, $z^\text{img}_i$ indicates the image feature of target image $x_i$, and $z^\text{text}_j$ represents the $j$-th element of the text feature set $Z^\text{text}$.
Conversely, as we lack information on the target-domain class name, generating negative prototypes via class names is impractical. Therefore, to locate negative prototypes for a known class $c$, we employ K-means \cite{K-means} to cluster all target image features and generate an image prototype (cluster centroid) set $V^\text{img}=\{v_k^\text{img}\}_{k=1}^K$, where $K$ is the K-means hyperparameter. The formal definition is as follows:
\begin{equation}
\{v_k^\text{img}\}_{k=1}^K = \text{K-means}(\{z^\text{img}_i\}_{i=1}^N),
\label{eq:generate_image_prototype}
\end{equation}
where $N$ is the number of target samples.
\textcolor{black}{We will later discuss optimal $K$ value determination.} The $K$ value sensitivity analysis can be found in the experiment. After clustering, there are $K$ image prototypes. We assume there are one positive image prototype $p^c$ and $K-1$ negative image prototypes $\{n^c_k\}_{k=1}^{K-1}$ for a known class $c$. $p^c$ and $\{n^c_k\}_{k=1}^{K-1}$ for a known class $c$ are determined by:
\begin{equation}
\begin{cases}
p^c = v_j^\text{img}, \; \text{if} \; \cos(z^\text{text}_c,v_j^\text{img})=\max\{\cos(z^\text{text}_c,v_k^\text{img})\}_{k=1}^K \\
\{n^c_k\}_{k=1}^{K-1} = \{v_k^\text{img}\}_{k=1}^K/\{v_j^\text{img}\}.
\end{cases}
\label{eq:positive_negative_image_prototypes}
\end{equation}
In this paper, we use the text feature $z^\text{text}_c$ generated by the text encoder instead of the positive image prototype $p^c$ as the positive prototype, and we employ the negative image prototypes $\{n^c_k\}_{k=1}^{K-1}$ as the negative prototypes. 
According to the information bottleneck theory \cite{information_bottleneck}, we need to maximize the objective function 
\begin{equation}
R_\text{IB}(\omega) = I(Z,Y;\omega) - \beta I(Z,X;\omega),
\end{equation}
where $R_\text{IB}$ is the information bottleneck, $X$ is the input set, $Y$ is the output set, and $I$ denotes the mutual information of input/output sets with the feature set $Z$. Here, the parameters $\omega$ of encoders in VLMs are held constant, causing the value of $R_\text{IB}$ to be contingent upon the features within $Z$. The image features, $Z^\text{img}$, derived from the target images $X$, inherently encapsulate domain-specific information. In contrast, the text features, $Z^\text{text}$, exhibit similarities to image prototypes but lack domain-related information, i.e., $I(Z^\text{img},X;\omega) > I(Z^\text{text},X;\omega)$, which further implies $I(V^\text{img},X;\omega) > I(Z^\text{text},X;\omega)$. For another thing, the image prototypes $V^\text{img}$ resulting from a limited set of images may not adequately capture the essence of a class. E.g., \textcolor{black}{the distribution of images for subclasses such as pony and zebra within the \texttt{horse} class in DomainNet is imbalanced,} rendering the image prototype inadequate in representing the \texttt{horse} class. Consequently, we deduce that $ I(Z^\text{text},Y;\omega) > I(V^\text{img},Y;\omega)$. This leads us to conclude that:
\begin{equation}
R_\text{IB}(Z^\text{text}) > R_\text{IB}(V^\text{img}).
\label{eq:ib_comparison}
\end{equation}
Furthermore, we observe that the quality of the positive image prototype $p^c$ heavily relies on the empirical determination of appropriate hyperparameters $K$ for clustering. If the hyperparameter $K$ is set smaller than the oracle $K$, it poses the risk of merging two distinct classes into a single positive image prototype. If $K$ is larger than the oracle $K$, the positive image prototype may represent a subclass concept rather than the intended class concept. On the other hand, the text feature $z^\text{text}_c$, derived from text, remains unaffected by $K$. Thus, it becomes evident that utilizing the text feature $z^\text{text}_c \in Z^\text{text}$ as the positive prototype stands as a more rational choice than employing the image prototype $p^c \in V^\text{img}$.

\textbf{\textcolor{black}{Self-Training.}} Given the positive prototype $z^\text{text}_c$ and the set of negative prototypes $\{n^c_k\}_{k=1}^{K-1}$, the ACTP module generates a pseudo label $\hat{y}_i$ for a target sample $x_i$ to implement self-training. The formal definition is as follows:
\begin{equation}
\hat{y}_i=
\begin{cases}
\text{one-hot($c$)},  &  \text{if} \; \exists p(y=c|x_i)\ge \max\{\cos(z^\text{text}_c,n^c_k)\}_{k=1}^{K-1} \\
\texttt{unknown}, & \text{else}
\end{cases}
\label{eq:generate_pseudo_label}
\end{equation}
where $p(y=c|x_i)$ is the probability in \cref{eq:pred_probability}. As shown in \cref{Fig:model_diagram}, for a target sample $x_i$ with pseudo label \verb+unknown+, ACTP assigns a uniform encoding $\hat{y}_i=[1/|C^s|,...,1/|C^s|] \in \mathbb{R}^{1 \times |C^s|}$ to increase the uncertainty for $x_i$. Hence, the model can distinguish common and unknown class samples at the inference phase.

\textbf{Image Loss and Text Loss.} The image cross-entropy loss with pseudo labels $\hat{y}_i$ is defined as follows:
\begin{equation}
\mathcal{L}^{\text{img}} = -\frac{1}{N} \sum_{i=1}^{N}\hat{y}_i\log(\sigma(h_\theta(z^\text{img}_i))),
\label{eq:image_ce_loss}
\end{equation}
where $\sigma$ and $h_\theta$ denote the softmax function and the closed-set classifier, respectively. For another thing, given a text feature derived from a known (source) class name such as \texttt{horse}, its ground truth label can be determined. The text cross-entropy loss with ground truth labels $y_c$ is defined as follows:
\begin{equation}
\mathcal{L}^{\text{text}} = -\frac{1}{|C^s|}\displaystyle\sum_{\substack{c=1}}^{\substack{|C^s|}}y_c\log(\sigma(h_\theta(z^\text{text}_c))).
\label{eq:text_ce_loss}
\end{equation}

\textbf{Optimal $K$ Determination.} To determine the optimal $K$ for K-means, various methods have been introduced in \cite{Silhouettes,dendrite,A_Cluster_Separation_Measure}. Analysis of the effects of these methods on COCA can be found in the appendix. \textcolor{black}{In summary, COCA performs stably across these methods.} In this paper, we employ the Silhouette metric \cite{Silhouettes} to estimate the value of $K$. The introduction to the Silhouette score can be found in the appendix. We consider the potential values of $K$ as $[1/3|C^s|, 1/2|C^s|, |C^s|, 2|C^s|, 3|C^s|]$ following \cite{GLC}. The model selects the optimal $K$ from this list using the Silhouette value. Since we freeze the image and text encoders, the image and text features do not change. Hence, we estimate the $K$ value only at the first epoch.

%% file: eccv24/3.2_mieci.tex
In order to encourage the closed-set classifier to exploit the context information in images and enhance the mutual information, we introduce the mutual information enhancement by context information (MIECI) module. 

\textbf{Masked Image and EMA Teacher.} We use a patch mask $\mathsf{M}$ \cite{MIC}, which is randomly sampled from a uniform distribution, to mask out target samples:
\begin{equation}
\mathsf{M_{mp+1:(m+1)p,np+1:(n+1)p}=\mathbbm{1}_{(v>r)}} \; \text{with} \; \mathsf{v\sim U(0,1)},
\label{eq:generate_patch_mask}
\end{equation}
where $\mathsf{p}$ represents the patch size, $\mathsf{r}$ is the mask ratio, and $\mathsf{m\in\{0,...,s/p-1\}}$, $\mathsf{n\in\{0,...,s/p-1\}}$ are the patch indices, $\mathsf{s}$ is the input image size. Masked target image $x^\mathsf{M}_i$ is obtained by performing element-wise multiplication of the mask $\mathsf{M}$ and target image $x_i$:
\begin{equation}
x^\mathsf{M}_i = \mathsf{M} \odot x_i.
\label{eq:generate_masked_target_image}
\end{equation}
We use the exponential moving average (EMA) teacher \cite{EMA_teacher} as the classifier $h_\gamma$. Its weights are updated via the weights of $h_\theta$ with a smoothing factor $\alpha$:
\begin{equation}
\gamma_{t+1} \leftarrow \alpha\gamma_{t} + (1-\alpha)\theta_t,
\label{eq:update_teacher_classifier}
\end{equation}
where $t$ denotes a training step.
Unmasked target image $x_i$ is fed to the teacher classifier $h_\gamma$ and masked target image $x^\mathsf{M}_i$ to the classifier $h_\theta$. The probabilities $q_i$ generated by the teacher classifier $h_\gamma$ for $x_i$ are utilized
as soft labels to construct the mask loss $\mathcal{L}^\text{mask}$ to guide the classifier $h_\theta$. The formal definition of $q_i$ is:
\begin{equation}
q_i = \sigma(h_\gamma(G^{\text{img}}(x_i))).
\end{equation}
Additionally, since classifiers $h_\gamma,h_\theta$ are closed-set, we need to select the target samples belonging to known classes. In this paper, we use the pseudo labels $\hat{y}_i$ generated by the ACTP module to assist in selecting known class samples. 

\textbf{Mask Loss}. The mask loss $\mathcal{L}^\text{mask}$ in the MIECI module is defined as:
\begin{equation}
\mathcal{L}^\text{mask} = \mathbb{E}(-\mathbbm{1}_{(\hat{y}_i\in C^s)}q_i\log(\sigma(h_\theta(G^{\text{img}}(x^\mathsf{M}_i))))),
\label{eq:cal_mask_loss}
\end{equation}
where $\mathbb{E}$ indicates expectation. The mask loss $\mathcal{L}^\text{mask}$ encourages the classifier $h_\theta$ to produce a similar probability distribution, akin to the EMA teacher $h_\gamma$, when provided with a masked known class image $x^\mathsf{M}_i$. Essentially, it drives $h_\theta$ to ignore subtly differing visual appearances of the same image. This strategy enhances the capacity of $h_\theta$ to effectively harness contextual cues. Further, $h_\theta$ is trained to disregard minor dissimilarities within a given common class, e.g., pony and zebra both belong to \texttt{horse} class in DomainNet or the same kind of object with various perspectives. This process enhances mutual information $I(Z^\text{img}, Y; h_\theta)$ and, in turn, improves classification accuracy for common class samples.\label{Dis:reason_mieci}

%% file: eccv24/3.3_model_optimization.tex
The overall training loss of our approach can be written as:
\begin{equation}
\mathcal{L}=\mathcal{L}^{\text{img}}+\mathcal{L}^{\text{text}}+\mathcal{L}^\text{mask}.
\end{equation}

\label{Dis:reason_adapt_decision_boundary}
\textbf{Decision Boundary Adaptation.} The general objective of DA is to establish a decision boundary by minimizing classification loss of source samples and explicitly design a loss term measuring domain divergence to make the fixed decision boundary suitable for target samples. This objective aims to transfer the source domain knowledge to the target domain. (1) In contrast to DA, SF-UniDA tasks do not allow direct access to source samples, making it impossible to estimate the source data distribution. As a result, explicitly designing a loss term for measuring domain divergence is not feasible. Consequently, in this work, we consider to adapt the decision boundary, i.e., the classifier. The core idea is if we properly adapt the decision boundary to target domains, the classification error will be low without the need for explicitly designing a domain shift loss term. (2) We contend that the knowledge of both the source and target domains has been encapsulated within the VLMs, given its pretraining on large datasets. \textcolor{black}{Therefore, the explicit design of a domain divergence measurement term for optimizing the image and text encoders to transfer source domain knowledge becomes redundant.} The focus, instead, should be shifted toward guiding the classifier to establish a more appropriate decision boundary. \textcolor{black}{Given the above two reasons, we propose this new paradigm as shown in \cref{Fig:Pipeline_comparison}, which concentrates on classifier optimization.} In this study, we exploit the \textcolor{black}{close feature distance} between image prototypes and text features to generate positive and negative prototypes, \textcolor{black}{thereby adapting the classifier's decision boundary to new target domains. In the experiment, we demonstrate that our new paradigm is applicable not only to VLM-powered few-shot learners but also to zero-shot classifiers such as single linear layer \cite{CLIP} or the adapter module \cite{gao2021clipadapter}, suggesting that the image and text encoders within the VLMs have encapsulated knowledge of both the source and target domains.}

%% file: eccv24/3.4_inference.tex
Since $h_\theta$ is a closed-set classifier, we utilize the normalized Shanon Entropy \cite{A_mathematical_theory_of_communication} to measure the uncertainty of target samples $U(x_i)\in[0,1]$ to distinguish common and unknown class samples: 
\begin{equation}
U(x_i)=-\frac{1}{\log |C^s|}\sigma(h_\theta(z^\text{img}_i))\log (\sigma(h_\theta(z^\text{img}_i))).
\end{equation}
At the inference phase, we separate common and unknown class samples as:
\begin{equation}
    y(x_i)=
    \begin{cases}
    \text{argmax}(h_\theta(z^\text{img}_i)), & \text{if} \; U(x_i) < \tau \\
    \texttt{unknown}. & \text{else}.
    \end{cases}
    \label{eq:ditinguish_known_unknown}
\end{equation}
The inference phase is depicted in \cref{fig:test_phase}. We set $\tau=0.55$ for all our experiments. Its sensitivity analysis can be found in the appendix.

%% file: eccv24/4.1_datasets_and_evaluation_metric.tex
\begin{figure}[t]
\begin{minipage}{0.37\textwidth}
\includegraphics[width=\textwidth]{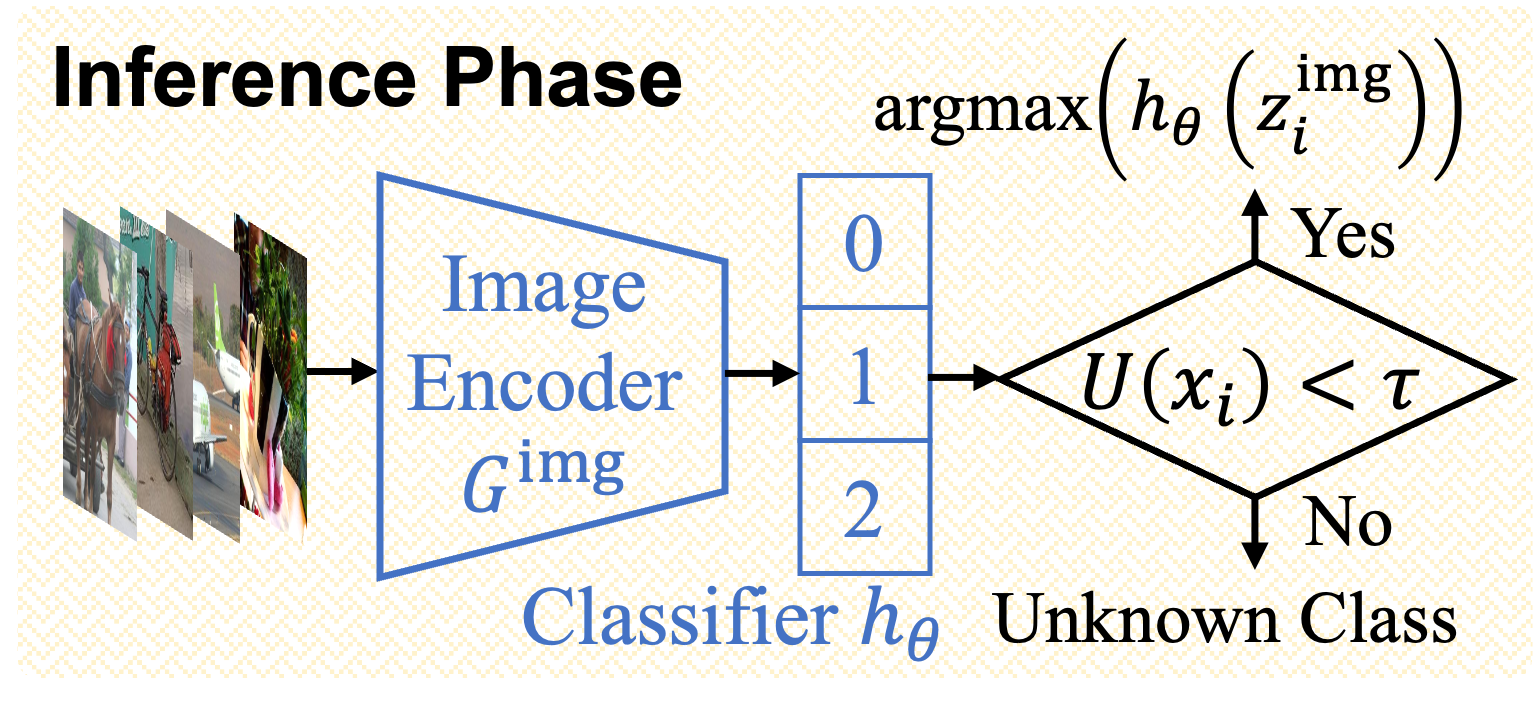}
\caption{COCA at the inference phase.}
\label{fig:test_phase}
\end{minipage}
\hfill
\begin{minipage}{0.63\textwidth}
\centering
\captionof{table}{Details regarding the split of classes and the average number of source samples per class provided to methods.}
\resizebox{\textwidth}{!}{
\begin{tabular}{c|ccc|cc}
\toprule
\multirow{2}{*}{Dataset} & \multicolumn{3}{c|}{Class Split ($C/ \bar{C}^s/ \bar{C}^t$)} & \multicolumn{2}{c}{Number of Source Samples Per Class}  \\ 
\cmidrule{2-6} & OPDA  & OSDA  & PDA & Others & Ours  \\ 
\midrule
OfficeHome & 10/5/50 & 25/0/40 & 25/40/0 & 60 (full set) & [0, 1, 4, 16] \\
VisDA-2017 & 6/3/3 & 6/0/6  & 6/6/0 & 12,700 (full set) & [0, 1, 4, 16]  \\
DomainNet & 150/50/145   & -       & - & 304 (full set) & [0, 1, 4, 16] \\ 
\bottomrule
\end{tabular}
}
\label{tab:class_split_source_comparison}
\end{minipage}
\end{figure}

\textbf{Dataset}. We utilize the following three public benchmarks: \textbf{OfficeHome} \cite{OfficeHome}, \textbf{VisDA-2017} \cite{VisDa}, and \textbf{DomainNet} \cite{DonmaiNet}, to evaluate the effectiveness of our approach.
OfficeHome consists of four domains: Art (Ar), Clipart (Cl), Product (Pr), and Real-World (Rw).
Following preceding studies \cite{GATE,GLC}, our experiments were conducted on three domains of DomainNet: Painting (P), Real (R), and Sketch (S). We conducted evaluations of our method in OPDA, OSDA, and PDA. A summary of class splits and the comparison of the number of source samples per class on average provided to different methods are detailed in \cref{tab:class_split_source_comparison}. The classes are separated according to their alphabetical order. While other methods are fully trained on source samples, our approach was allotted merely few-shot source samples per class for source model training.

\textbf{Implementation.} The source model is based on CLIP(ViT-16/B). At the target domain adaptation phase, we applied the AdamW \cite{AdamW} optimizer, configured with beta values of (0.9, 0.999), an epsilon of 1e-08, and a weight decay of 0.01. The batch size is 64 for all benchmarks. The learning rate was adjusted according to the sample number of target domains, resulting in rates of 1e-3 for OfficeHome, 1e-4 for VisDA-2017, and 1e-5 for DomainNet. The MIECI module uses mask ratio $\mathsf{v}=0.5$, a patch size $\mathsf{p}=16$, a smooth factor $\alpha=0.999$ suggested by \cite{EMA_teacher}, and color augmentation following the parameters suggested by \cite{MIC}. All of our experiments are conducted using an RTX-4090 GPU and PyTorch-2.0.1.

\textbf{Evaluation Metric.} We use the same evaluation metric as previous works \cite{GATE,GLC} for a fair comparison. Specifically, we report $HOS$ \cite{ROS}, also referred to as H-score \cite{CMU}, to evaluate the model performance in OPDA and OSDA scenarios. $HOS$ is a harmonic mean of the average class accuracy over the known classes $OS^*$ and the accuracy of the unknown class $UNK$: $HOS = 2 \times \frac{OS^* \times UNK}{OS^* + UNK}$.
In the PDA scenario, we use classification accuracy over all target samples.

\begin{table*}[t]
\centering
  \caption{HOS (\%) comparison in \textbf{OPDA} on OfficeHome. \textbf{SF} denotes support for source-free learning, while \textbf{FSL} indicates support for few-shot learning on the source domain. Best in \textbf{bold}.}
  \label{table:reluts_officehome_opda}
  \resizebox{\textwidth}{!}{
  \begin{tabular}{l|c|cc|ccccccccccccc}
  \toprule
  Method & & SF & FSL & Ar$\rightarrow$Cl & Ar$\rightarrow$Pr & Ar$\rightarrow$Rw & Cl$\rightarrow$Ar & Cl$\rightarrow$Pr & Cl$\rightarrow$Rw & Pr$\rightarrow$Ar & Pr$\rightarrow$Cl & Pr$\rightarrow$Rw & Rw$\rightarrow$Ar & Rw$\rightarrow$Cl & Rw$\rightarrow$Pr & \textbf{Avg}\\ 
  \midrule
  OSBP~\cite{OSBP}  & \multirow{9}{*}{\rotatebox{90}{ResNet50}} & \xmark & \xmark & 39.6 & 45.1 & 46.2 & 45.7 & 45.2 & 46.8 & 45.3 & 40.5 & 45.8 & 45.1 & 41.6 & 46.9 & 44.5\\
  DCC~\cite{DCC} &  & \xmark & \xmark & 58.0 & 54.1 & 58.0 & 74.6 & 70.6 & 77.5 & 64.3 & 73.6 & 74.9 & 81.0 & 75.1 & 80.4 & 70.2 \\ 
  DCC+SPA~\cite{SPA} & & \xmark & \xmark & 59.3 & 79.5 & 81.5 & 74.7 & 71.7 & 82.0 & 68.0 & 74.7 & 75.8 & 74.5 & 75.8 & 81.3 & 74.9 \\
  OVANet~\cite{OVANet}  & & \xmark & \xmark & 62.8   & 75.5  & 78.6  & 70.7   & 68.8  & 75.0 & 71.3 & 58.6 & 80.5 &  76.1 & 64.1 & 78.9   &  71.8\\ 
  GATE~\cite{GATE} & & \xmark & \xmark & 63.8 & 70.5& 75.8& 66.4 & 67.9 & 71.7 & 67.3 & 61.5 & 76.0 & 70.4 & 61.8 & 75.1 & 69.0 \\
  UniOT~\cite{UniOT} & & \xmark & \xmark & 67.3 & 80.5 & 86.0 & 73.5 & 77.3 & 84.3 & 75.5 & 63.3 & 86.0 & 77.8 & 65.4 & 81.9 & 76.6 \\
  UMAD~\cite{UMAD} & & \cmark & \xmark & 61.1 & 76.3 & 82.7 & 70.7 & 67.7 & 75.7 & 64.4 & 55.7 & 76.3 & 73.2 & 60.4 & 77.2 & 70.1 \\
  GLC~\cite{GLC} & & \cmark & \xmark & 64.3 & 78.2 & 89.8 & 63.1 & 81.7& 89.1 & 77.6 & 54.2 & 88.9 & 80.7 & 54.2 & 85.9 & 75.6\\ 
  CoDE~\cite{CoDE} & & \cmark & \xmark & 65.8 & 77.4 & 87.6 & 73.5 & 69.1 & 80.2 & 77.6 & 62.0 & 84.1 & 84.3 & 68.4 & 83.1 & 76.1 \\
  \midrule
  OVANet~\cite{OVANet} & \multirow{4}{*}{\rotatebox{90}{ViT}} & \xmark & \xmark & 55.8 & 77.6 & 86.2 & 72.6 & 69.9 & 81.1 & 76.3 & 47.4 & 84.7 & 79.9 & 53.0 & 79.6 & 72.0 \\
  UniOT~\cite{UniOT} & & \xmark & \xmark & 63.8 & 88.2 & 90.2 & 75.0 & 81.0 & 84.6 & 78.9 & 61.3 & 87.6 & 82.4 & 63.7 & 88.3 & 78.4 \\
 UniAM~\cite{UniAM} & & \xmark & \xmark & 72.0 & 87.1 & 90.7 & 80.3 & 82.4 & 79.8 & 85.0 & 68.4 & 89.0 & 85.4 & 72.1 & 86.1 & 81.7 \\
 GLC~\cite{GLC} & & \cmark & \xmark & 68.5 & \textbf{89.8} & \textbf{91.0} & 82.4 & \textbf{88.1} & \textbf{89.4} & 82.1 & 69.7 & 88.2 & 82.4 & 70.9 & 88.9 & 82.6 \\
  \midrule
  DCC~\cite{DCC} & \multirow{3}{*}{\rotatebox{90}{CLIP}}  & \xmark & \xmark & 62.6 & 88.7 & 87.4 & 63.3 & 68.5 & 79.3 & 67.9 & 63.8 & 82.4 & 70.7 & 69.8 & 87.5 & 74.4 \\
  OVANet~\cite{OVANet} & & \xmark & \xmark & 71.0 & 85.4 & 86.8 & 79.4 & 78.8 & 85.4 & 76.8 & 64.8 & \textbf{89.1} & 83.2 & 70.3 & 84.4 & 79.6 \\
  GLC~\cite{GLC} & & \cmark & \xmark & 79.4 & 88.9 & 90.8 & 76.3 & 84.7 & 89.0 & 71.5 & 72.9 & 85.7 & 78.2 & 79.4 & \textbf{90.0} & 82.2 \\
  \midrule
  Source Model (16-shot) \cite{cross-modal_adaptation} &  & \xmark & \cmark & 43.5 & 31.6 & 39.5 & 52.0 & 33.9 & 43.1 & 55.7 & 51.9 & 44.0 & 49.9 & 54.4 & 27.6 & 43.9 \\
  \rowcolor{mygray} 
  + COCA-w-$p^c$ & & \cmark & \cmark & 83.6 & 85.2 & 87.2 & 81.9 & 84.1 & 86.7 & 78.0 & 82.4 & 86.7 & 82.7 & 82.8 & 84.1 & 83.8\\
  \rowcolor{mygray} 
  + COCA & & \cmark & \cmark & \textbf{83.7} & 86.6 & 89.0 & \textbf{88.0} & 86.5 & 88.9 & \textbf{88.2} & \textbf{83.9} & 88.8 & \textbf{88.3} & \textbf{83.7} & 86.6 & \textbf{86.9}\\
    \bottomrule
  \end{tabular}
  }
\end{table*}

%% file: eccv24/4.2_comparison_with_sota.tex
\label{subsec:comparison_with_sota}
To assess the effectiveness of our method, we conducted extensive experiments across three category-shift scenarios: OPDA, OSDA, and PDA. \textcolor{black}{In \cref{subsec:comparison_with_sota} and \cref{subsec:further_evaluation}, the source model is the cross-modal linear probing model \cite{cross-modal_adaptation}.} \textbf{COCA-w-$p^c$} represents that we utilize the image prototype $p^c$ instead of the text feature $z^\text{text}_c$ as the positive prototype. We also included the results from DCC, OVANet, and GLC models with the ViT-B/16 \cite{ViT} pre-trained on ImageNet \cite{ImageNet} and the CLIP(ViT-B/16), to ensure a more fair comparison.

\begin{table*}[t] 
\centering
  \caption{HOS (\%) in \textbf{OPDA} on DomainNet and VisDA-2017.}
  \label{table:reluts_domainnet_visda_opda}
  \resizebox{0.7\textwidth}{!}{
  \begin{tabular}{l|c|cc|ccccccc|c}
  \toprule
  \multirow{2}{*}{Method} & & \multirow{2}{*}{SF} & \multirow{2}{*}{FSL} & \multicolumn{7}{c|}{DomainNet} & \multirow{2}{*}{VisDA} \\ \cmidrule{5-11}
  & &  &  & P$\rightarrow$R & P$\rightarrow$S & R$\rightarrow$P & R$\rightarrow$S & S$\rightarrow$P & S$\rightarrow$R & \textbf{Avg} &\\ 
  \midrule
  OSBP~\cite{OSBP}  & \multirow{9}{*}{\rotatebox{90}{ResNet50}} & \xmark & \xmark & 33.6 & 30.6 & 33.0 & 30.6 & 30.5 & 33.7 & 32.0 & 27.3\\
  DCC~\cite{DCC} &  & \xmark & \xmark & 56.9 & 43.7 & 50.3 & 43.3 & 44.9 & 56.2 & 49.2 & 43.0\\ 
  DCC+SPA~\cite{SPA} & & \xmark & \xmark & 59.1 & 52.7 & 47.6 & 45.4 & 46.9 & 56.7 & 51.4 & -\\
  OVANet~\cite{OVANet}  & & \xmark & \xmark & 56.0 & 47.1 & 51.7 & 44.9 & 47.4 & 57.2 & 50.7 & 53.1\\ 
  GATE~\cite{GATE} & & \xmark & \xmark & 57.4 & 48.7 & 52.8 & 47.6 & 49.5 & 56.3 & 52.1 & 56.4\\
  UniOT~\cite{UniOT} & & \xmark & \xmark & 59.3 & 51.8 & 47.8 & 48.3 & 46.8 & 58.3 & 52.0 & 57.3\\
  UMAD~\cite{UMAD} & & \cmark & \xmark & 59.0 & 44.3 & 50.1 & 42.1 & 32.0 & 55.3 & 47.1 & 58.3\\
  GLC~\cite{GLC} & & \cmark & \xmark & 63.3 & 50.5 & 54.9 & 50.9 & 49.6 & 61.3 & 55.1 & 73.1\\ 
  CoDE~\cite{CoDE} & & \cmark & \xmark & 62.7 & 47.1 & 51.3 & 43.4 & 48.3 & 60.9 & 52.3 & 74.5\\
  \midrule
  OVANet~\cite{OVANet} & \multirow{4}{*}{\rotatebox{90}{ViT}} & \xmark & \xmark & 65.0 & 42.5 & 54.7 & 37.7 & 40.5 & 58.9 & 49.9 & 49.6\\ 
  UniOT~\cite{UniOT} & & \xmark & \xmark & 72.4 & 49.3 & 59.5 & 47.4 & 56.9 & 69.4 & 59.1 & 63.3 \\
  UniAM~\cite{UniAM} & & \xmark & \xmark & 73.9 & 52.3 & 60.9 & 51.4 & 60.0 & 70.7 & 61.5 & 65.2 \\
 GLC~\cite{GLC} & & \cmark & \xmark & 67.6 & 51.1 & 55.9 & 46.6 & 53.8 & 66.0 & 56.8 & 80.3\\
  \midrule
  DCC~\cite{DCC} & \multirow{3}{*}{\rotatebox{90}{CLIP}} & \xmark & \xmark & 61.1 & 38.8 & 51.8 & 49.3 & 49.1 & 60.3 & 52.2 & 61.2 \\
  OVANet~\cite{OVANet} & & \xmark & \xmark & 65.4 & 53.7 & 56.3 & 53.1 & 55.9 & 67.2 & 58.6 & 72.0 \\
  GLC~\cite{GLC} & & \cmark & \xmark & 74.4 & 63.4 & 60.0 & 62.9 & 52.0 & 74.3 & 64.5 & 77.6 \\
  \midrule
  Source Model (16-shot) \cite{cross-modal_adaptation} &  & \xmark & \cmark & 46.5 & 51.8 & 56.3 & 53.9 & 32.7 & 33.2 & 45.7 & 32.2\\
  \rowcolor{mygray} 
  + COCA-w-$p^c$ & & \cmark & \cmark & 78.9 & \textbf{69.2} & 68.5 & \textbf{69.0} & 67.6 & 77.6 & 71.8 & 79.8\\
  \rowcolor{mygray} 
  + COCA  & & \cmark & \cmark & \textbf{80.8} & \textbf{69.2} & \textbf{69.6} & \textbf{69.0} & \textbf{69.4} & \textbf{80.5} & \textbf{73.1} & \textbf{83.2}\\
    \bottomrule
  \end{tabular}
  }
\end{table*}

\textbf{Results for OPDA.} Our first experiment focuses on the most challenging setting, i.e., OPDA, where both source and target domains contain private categories. Results for OfficeHome are shown in \cref{table:reluts_officehome_opda}, and those for VisDA-2017 and DomainNet are summarized in \cref{table:reluts_domainnet_visda_opda}.
As presented in \cref{table:reluts_officehome_opda} and \cref{table:reluts_domainnet_visda_opda}, our approach outperforms all previous methods, despite these methods utilizing the full set of source samples for model training. 
The source model, i.e., cross-modal linear probing \cite{cross-modal_adaptation}, utilizes the uncertainty in \cref{eq:ditinguish_known_unknown} to distinguish common and unknown class samples. The source model exhibits subpar performance in the OPDA scenario due to its inability to distinguish between common and unknown class samples. 
In the most challenging benchmark, i.e., DomainNet, our approach surpasses all previous works by a dramatically wide margin. It implies the potential of COCA on large-scale DA datasets.

\begin{table*}[t] 
\centering
  \caption{HOS (\%) comparison in \textbf{OSDA} on OfficeHome and VisDA-2017.}
  \label{table:reluts_officehome_visda_osda}
  \resizebox{\textwidth}{!}{
  \begin{tabular}{l|c|cc|ccccccccccccc|c}
  \toprule
  \multirow{2}{*}{Method} &  & \multirow{2}{*}{SF} & \multirow{2}{*}{FSL} & \multicolumn{13}{c|}{OfficeHome} & \multirow{2}{*}{VisDA} \\ \cmidrule{5-17}
   & & & & Ar$\rightarrow$Cl & Ar$\rightarrow$Pr & Ar$\rightarrow$Rw & Cl$\rightarrow$Ar & Cl$\rightarrow$Pr & Cl$\rightarrow$Rw & Pr$\rightarrow$Ar & Pr$\rightarrow$Cl & Pr$\rightarrow$Rw & Rw$\rightarrow$Ar & Rw$\rightarrow$Cl & Rw$\rightarrow$Pr & \textbf{Avg} & \\
  \midrule
  OSBP~\cite{OSBP}  & \multirow{8}{*}{\rotatebox{90}{ResNet50}} & \xmark & \xmark & 55.1 & 65.2 & 72.9 & 64.3 & 64.7 & 70.6 & 63.2 & 53.2 & 73.9 & 66.7 & 64.5 & 72.3 & 64.7 & 52.3\\
  $\text{STA}_\text{max}$~\cite{STA} &  & \xmark & \xmark & 55.8 & 54.0 & 68.3 & 57.4 & 60.4 & 66.8 & 61.9 & 53.2 & 69.5 & 67.1 & 54.5 & 64.5 & 61.1 & 65.0 \\
  DCC~\cite{DCC} &  & \xmark & \xmark & 56.1 & 67.5 & 66.7 & 49.6 & 66.5 & 64.0 & 55.8 & 53.0 & 70.5 & 61.6 & 57.2 & 71.9 & 61.7 & 59.6 \\ 
  OVANet~\cite{OVANet}  & & \xmark & \xmark & 58.9 & 66.0 & 70.4 & 62.2 & 65.7 & 67.8 & 60.0 & 52.6 & 69.7 & 68.2 & 59.1 & 67.6 & 64.0 & 66.1\\ 
  UADAL~\cite{UADAL} & & \xmark & \xmark & \textbf{76.9} & 56.6 & 63.0 & 70.8 & 77.4 & 63.2 & 72.1 & \textbf{76.8} & 60.6 & 73.4 & 64.2 & 69.5 & 68.7 & 61.3\\
  GATE~\cite{GATE} & & \xmark & \xmark & 63.8 & 70.5 & 75.8 & 66.4 & 67.9 & 71.7 & 67.3 & 61.5 & 76.0 & 70.4 & 61.8 & 75.1 & 69.0 & 70.8 \\
  UMAD~\cite{UMAD} & & \cmark & \xmark & 59.2 & 71.8 & 76.6 & 63.5 & 69.0 & 71.9 & 62.5 & 54.6 & 72.8 & 66.5 & 57.9 & 70.7 & 66.4 & 66.8  \\
  GLC~\cite{GLC} & & \cmark & \xmark & 65.3 & 74.2 & 79.0 & 60.4 & 71.6 & 74.7 & 63.7 & 63.2 & 75.8 & 67.1 & 64.3 & 77.8 & 69.8 & 72.5 \\ 
  \midrule
  OVANet~\cite{OVANet} & \multirow{2}{*}{\rotatebox{90}{ViT}} & \xmark & \xmark & 55.5 & 71.1 & 76.9 & 64.6 & 67.4 & 75.1 & 64.4 & 47.2 & 76.7 & 71.4 & 53.5 & 70.5 & 66.2 & 57.4\\
 GLC~\cite{GLC} & & \cmark & \xmark & 68.4 & 81.7 & \textbf{84.5} & 76.0 & 82.4 & \textbf{83.8} & 69.9 & 59.6 & \textbf{84.6} & 73.3 & 66.8 & 83.9 & 76.2 & 81.6  \\
  \midrule
  DCC~\cite{DCC} & \multirow{3}{*}{\rotatebox{90}{CLIP}} & \xmark & \xmark & 62.9 & 73.3 & 78.4 & 49.8 & 69.2 & 75.0 & 59.3 & 61.5 & 80.9 & 68.1 & 62.5 & 80.0 & 68.4 & 66.2 \\
  OVANet~\cite{OVANet} &  & \xmark & \xmark & 65.0 & 73.6 & 77.5 & 71.1 & 73.9 & 79.0 & 65.7 & 54.6 & 80.3 & 73.2 & 61.1 & 77.1 & 71.0 & 70.7\\
  GLC~\cite{GLC} &  & \cmark & \xmark & 72.1 & 79.7 & 83.3 & 55.5 & 81.3 & 77.9 & 52.1 & 65.9 & 78.2 & 69.0 & 71.3 & 83.9 & 72.5 & 83.4 \\
  \midrule
  Source Model (16-shot) \cite{cross-modal_adaptation} &  & \xmark & \cmark & 33.3 & 13.5 & 17.8 & 36.0 & 18.5 & 23.0 & 37.4 & 48.6 & 28.1 & 18.3 & 11.7 & 8.8 & 24.6 & 55.6 \\
  \rowcolor{mygray} 
  + COCA-w-$p^c$ & & \cmark & \cmark & 68.3 & \textbf{85.5} & 78.9 & 73.4 & \textbf{86.8} & 79.0 & 73.6 & 67.9 & 78.5 & 75.4 & 73.0 & \textbf{85.7} & 77.2 & 70.7\\
  \rowcolor{mygray} 
  + COCA & & \cmark & \cmark & 75.6 & 84.5 & 82.5 & \textbf{79.7} & 84.3 & 82.5 & \textbf{79.6} & 74.5 & 82.5 & \textbf{80.0} & \textbf{75.7} & 84.4 & \textbf{80.5} & \textbf{86.3}\\
    \bottomrule
  \end{tabular}
  }
\end{table*}

\textbf{Results for OSDA.} Subsequent experiments are conducted on OSDA, wherein the target domain exclusively contains categories absent in the source domain. The corresponding results for OfficeHome and VisDA are shown in \cref{table:reluts_officehome_visda_osda}. As evidenced in \cref{table:reluts_officehome_visda_osda}, our method surpasses all prior models, despite the source model training based on few-shot source samples. These results indicate that our proposed plug-and-play approach can aid the closed-set few-shot learner in differentiating between common and unknown class samples within the target domain, consequently lowering the total labeling cost.

\begin{table*}[t] 
\centering
  \caption{Accuracy (\%) comparison in \textbf{PDA} on OfficeHome and VisDA-2017.}
  \label{table:reluts_officehome_visda_pda}
  \resizebox{\textwidth}{!}{
  \begin{tabular}{l|c|cc|ccccccccccccc|c}
  \toprule
  \multirow{2}{*}{Method} &  & \multirow{2}{*}{SF} & \multirow{2}{*}{FSL} & \multicolumn{13}{c|}{OfficeHome} & \multirow{2}{*}{VisDA} \\ \cmidrule{5-17}
   & & & & Ar$\rightarrow$Cl & Ar$\rightarrow$Pr & Ar$\rightarrow$Rw & Cl$\rightarrow$Ar & Cl$\rightarrow$Pr & Cl$\rightarrow$Rw & Pr$\rightarrow$Ar & Pr$\rightarrow$Cl & Pr$\rightarrow$Rw & Rw$\rightarrow$Ar & Rw$\rightarrow$Cl & Rw$\rightarrow$Pr & \textbf{Avg} & \\ 
  \midrule
  ETN~\cite{ETN} & \multirow{8}{*}{\rotatebox{90}{ResNet50}} & \xmark & \xmark & 59.2 & 77.0 & 79.5 & 62.9 & 65.7 & 75.0 & 68.3 & 55.4 & 84.4 & 75.7 & 57.7 & 84.5 & 70.4 & 59.8 \\
  BA3US~\cite{BA3US} & & \xmark & \xmark & 60.6 & 83.2 & 88.4 & 71.8 & 72.8 & 83.4 & 75.5 & 61.6 & 86.5 & 79.3 & 62.8 & 86.1 & 76.0 & 54.9 \\
  DANCE~\cite{DANCE} & & \xmark & \xmark & 53.6 & 73.2 & 84.9 & 70.8 & 67.3 & 82.6 & 70.0 & 50.9 & 84.8 & 77.0 & 55.9 & 81.8 & 71.1 & 73.7\\
  DCC~\cite{DCC} & & \xmark & \xmark & 54.2 & 47.5 & 57.5 & 83.8 & 71.6 & 86.2 & 63.7 & 65.0 & 75.2 & 85.5 & \textbf{78.2} & 82.6 & 70.9 & 72.4 \\
  OVANet~\cite{OVANet} & & \xmark & \xmark & 34.1 & 54.6 & 72.1 & 42.4 & 47.3 & 55.9 & 38.2 & 26.2 & 61.7 & 56.7 & 35.8 & 68.9 & 49.5 & 34.3 \\
  GATE~\cite{GATE} & & \xmark & \xmark & 55.8 & 75.9 & 85.3 & 73.6 & 70.2 & 83.0 & 72.1 & 59.5 & 84.7 & 79.6 & 63.9 & 83.8 & 74.0 & 75.6  \\
  SHOT-P~\cite{SHOT} & & \cmark & \xmark & 64.7 & 85.1 & 90.1 & 75.1 & 73.9 & 84.2 & 76.4 & 64.1 & 90.3 & 80.7 & 63.3 & 85.5 & 77.8 & 74.2 \\
  GLC~\cite{GLC} & & \cmark & \xmark & 55.9 & 79.0 & 87.5 & 72.5 & 71.8 & 82.7 & 74.9 & 41.7 & 82.4 & 77.3 & 60.4 & 84.3 & 72.5 & 76.2 \\ 
  \midrule
  OVANet~\cite{OVANet} & \multirow{2}{*}{\rotatebox{90}{ViT}} & \xmark & \xmark & 34.7 & 61.3 & 76.8 & 49.8 & 51.5 & 60.6 & 43.9 & 25.9 & 70.1 & 63.1 & 32.5 & 72.9 & 53.6 & 32.0 \\
  GLC~\cite{GLC} & & \cmark & \xmark & 63.2 & 80.7 & 86.5 & 76.0 & 77.9 & 84.1 & 74.5 & 56.8 & 84.7 & 79.8 & 57.4 & 83.0 & 75.4 & 84.0  \\
  \midrule
  DCC~\cite{DCC} & \multirow{3}{*}{\rotatebox{90}{CLIP}} & \xmark & \xmark & 59.4 & 78.8 & 83.2 & 62.0 & 78.6 & 79.3 & 64.2 & 44.4 & 82.9 & 76.5 & 70.7 & 84.6 & 72.1 & 79.8 \\
  OVANet~\cite{OVANet} & & \xmark & \xmark & 39.2 & 55.7 & 72.8 & 61.9 & 63.4 & 71.6 & 47.6 & 30.0 & 73.2 & 61.0 & 40.1 & 70.5 & 57.3 & 41.4 \\
  GLC~\cite{GLC} & & \cmark & \xmark & \textbf{77.8} & 82.8 & 89.5 & 68.7 & 81.8 & 86.4 & 74.3 & \textbf{75.3} & 86.3 & 79.0 & 78.1 & \textbf{87.2} & 80.6 & 84.4 \\
  \midrule
  Source Model (16-shot) \cite{cross-modal_adaptation} &  & \xmark & \cmark & 71.6 & 84.9 & 88.4 & 80.7 & 81.2 & 87.8 & 79.7 & 73.3 & 87.6 & \textbf{85.7} & 74.1 & 86.1 & 81.8 & 85.6 \\
  \rowcolor{mygray} 
  + COCA & & \cmark & \cmark & 69.1 & \textbf{87.7} & \textbf{92.4} & \textbf{83.9} & \textbf{86.7} & \textbf{90.7} & \textbf{83.8} & 68.6 & \textbf{90.5} & 84.3 & 69.7 & \textbf{87.2} & \textbf{82.9} & \textbf{89.1} \\
    \bottomrule
  \end{tabular}
  }
\end{table*}

\textbf{Results for PDA.} Lastly, we evaluate the effectiveness of COCA on PDA, where the class set $C^t$ of the target domain is a subset of the source domain. The shown results in \cref{table:reluts_officehome_visda_pda} demonstrate that our proposed method surpasses previous approaches, even those \cite{ETN,BA3US} specifically designed for PDA.

%% file: eccv24/4.3_further_evaluations.tex
\begin{table*}[t]
\caption{HOS (\%) with respect to $K$ in \textbf{OPDA}. COCA and COCA-w-$p^c$ are based on the source model (16-shot)\cite{cross-modal_adaptation}.}
    \label{tab:HOS_comparison_various_K}
    \resizebox{\textwidth}{!}{
    \begin{tabular}{c|cccccc|cccccc|cccccc}
    \toprule
    \multirow{2}{*}{Model} & \multicolumn{6}{c|}{OfficeHome ($|C^t|=60$)} & \multicolumn{6}{c|}{VisDA-2017 ($|C^t|=9$)} & \multicolumn{6}{c}{DomainNet ($|C^t|=295$)} \\ \cmidrule{2-19}
    & $K$=8 & $K$=15 & $K$=30 & $K$=45 & $K$=60 & $K$=75 & $K$=5 & $K$=9 & $K$=18 & $K$=27 & $K$=36 & $K$=45 & $K$=100 & $K$=200 & $K$=400 & $K$=600 & $K$=800 & $K$=1000 \\
    \midrule
    GLC (ViT) & 75.5 & 77.4 & 80.5 & 82.0 & 74.1 & 68.0 & 73.7 & 80.3 & 47.8 & 24.1 & 23.2 & 21.1 & 52.0 & 56.8 & 47.0 & 37.4 & 25.2 & 24.5 \\
    GLC (CLIP) & 79.3 & 80.7 & 82.5 & 81.4 & 78.8 & 76.5 & 62.6 & 77.6 & 53.0 & 10.2 & 9.7 & 9.6 & 62.0 & 64.5 & 53.8 & 46.6 & 41.1 & 46.3 \\
    COCA-w-$p^c$ & 70.7 & 73.0 & 78.8 & 83.8 & 84.6 & 82.8 & 78.8 & 79.8 & 63.4 & 44.8 & 36.2 & 22.4 & 66.5 & 70.0 & 71.8 & 72.3 & 72.0 & 72.0  \\
    COCA & \textbf{85.1} & \textbf{85.9} & \textbf{86.7} & \textbf{86.9} & \textbf{87.3} & \textbf{87.4} & \textbf{82.4} & \textbf{83.2} & \textbf{83.6} & \textbf{83.2} & \textbf{83.7} & \textbf{83.3} & \textbf{73.0} & \textbf{73.1} & \textbf{73.0} & \textbf{72.8} & \textbf{72.7} & \textbf{72.6} \\
    \midrule 
    \end{tabular}
    }
\end{table*}

\label{subsec:further_evaluation}
\textbf{Hyperparameter Sensitivity.} The comparison experiments of $K$ values are shown in \cref{tab:HOS_comparison_various_K}. COCA, exploiting textual prototypes, exhibits consistent performance across various $K$ values and is better and more stable than COCA-w-$p^c$ and the best-performed GLC \cite{GLC}, both utilizing image-based positive prototypes. COCA-w-$p^c$ and GLC are sensitive to the choice of $K$ values. In cases where $K$ values are inappropriate, the image-based positive prototypes may fail to accurately represent corresponding classes. Specifically, if $K$ values are smaller than the \textcolor{black}{oracle $K$}, distinct classes might erroneously merge into a single cluster. Conversely, if $K$ values are larger than the oracle $K$, different subclasses within the same class may be separated into distinct clusters, causing the image-based positive prototypes to represent subclasses instead of the intended classes. The experiment results and the analysis in \cref{eq:ib_comparison} indicate that \textbf{the textual prototypes are better suited as the positive prototypes}.

\begin{table*}[t]
\caption{HOS (\%) with respect to various source models in  \textbf{OPDA}.}
    \label{tab:HOS_comparison_various_source_model}
    \resizebox{\textwidth}{!}{
    \begin{tabular}{c|cc|cc|cc|cc|cc|cc|cc|cc}
    \toprule
    Source Model & \multicolumn{6}{c}{Linear Probe CLIP \cite{CLIP}} & \multicolumn{6}{|c}{CLIP-Adapter \cite{gao2021clipadapter}} & \multicolumn{4}{|c}{Cross-Modal Linear Probing \cite{cross-modal_adaptation}} \\
    \midrule
    & 0-shot & +COCA & 1-shot & +COCA & 16-shot & +COCA & 0-shot & +COCA & 1-shot & +COCA & 16-shot & +COCA & 1-shot & +COCA & 4-shot & +COCA \\
    \midrule
    OfficeHome & 61.1 & 86.9 & 57.7 & 86.7 & 51.7 & 86.7 & 73.7 & 87.1 & 71.4 & 87.0 & 58.5 & 86.9 & 52.9 & 86.9 & 57.7 & 87.0 \\
    VisDA-2017 & 59.2 & 83.0 & 41.1 & 83.6 & 44.0 & 83.4 & 71.0 & 83.0 & 46.0 & 83.7 & 43.4 & 83.5 & 41.1 & 82.9 & 37.3 & 82.8 \\
    DomainNet & 57.8 & 72.5 & 54.1 & 72.6 & 45.3 & 73.1 & 69.3 & 71.3 & 67.5 & 72.1 & 47.4 & 73.1 & 54.1 & 72.8 & 50.0 & 72.7  \\
    \bottomrule
    \end{tabular}
    }
\end{table*}

\textbf{Results on Varying Source Models.} \textcolor{black}{Results on varying source models in OPDA are shown in \cref{tab:HOS_comparison_various_source_model}.} With respect to the diverse tasks in OPDA, the outcomes reveal that the shots used in source model training do not significantly affect our method performance. We attribute this observation to the phenomenon of catastrophic forgetting. Specifically, since the few-shot source samples cannot be accessed at the target domain adaptation phase, the knowledge derived from a massive number of target samples with pseudo labels covers the knowledge from few-shot source samples.
In contrast to traditional UniDA methods, our approach's success is not predicated on knowledge extracted from a profusion of source samples. Instead, its effectiveness hinges upon the quality of \textcolor{black}{self-training}. Therefore, we choose the textual prototypes instead of the image prototypes as the positive ones since they demonstrate their advantages in SF-UniDA scenarios.
We posit that the knowledge learned from the massive number of labeled source samples is no longer a crucial factor for the success of VLMs in addressing the SF-UniDA challenge. It means the labeling cost for source samples in future works can be dramatically reduced compared with traditional SF-UniDA methods when utilizing VLMs.
Furthermore, the outcomes from diverse source models \cite{CLIP, gao2021clipadapter, cross-modal_adaptation} validate that our plug-and-play approach is compatible with multiple few-shot learning frameworks, ensuring consistent performance. \textcolor{black}{The results from "zero-shot linear probe CLIP + COCA" and "zero-shot CLIP-Adapter + COCA" indicate that \textbf{VLMs have encapsulated knowledge of both the source and target domains}. This suggests that we should \textbf{focus on classifier optimization to adapt the VLM-powered models to new target domains}.}

%% file: eccv24/5_conclusion.tex
In this paper, we present a novel plug-and-play method, called COCA, that endows VLM-powered few-shot learners with the unknown-aware ability to tackle the SF-UniDA challenge. Our paradigm shifts the focus to study classifier optimization for SF-UniDA, since we realize that VLMs have the ability to encapsulate knowledge of the source and target domains, to some extent. We hope that our method will inspire more SF-UniDA research.

%% file: eccv24/appendix/experiment_results.tex
\textcolor{black}{In this section, the source model is the cross-modal linear probing model \cite{cross-modal_adaptation}.}

\begin{figure}[t]
\centering
     \begin{subfigure}[b]{0.32\textwidth}
         \centering
         \includegraphics[width=\textwidth]{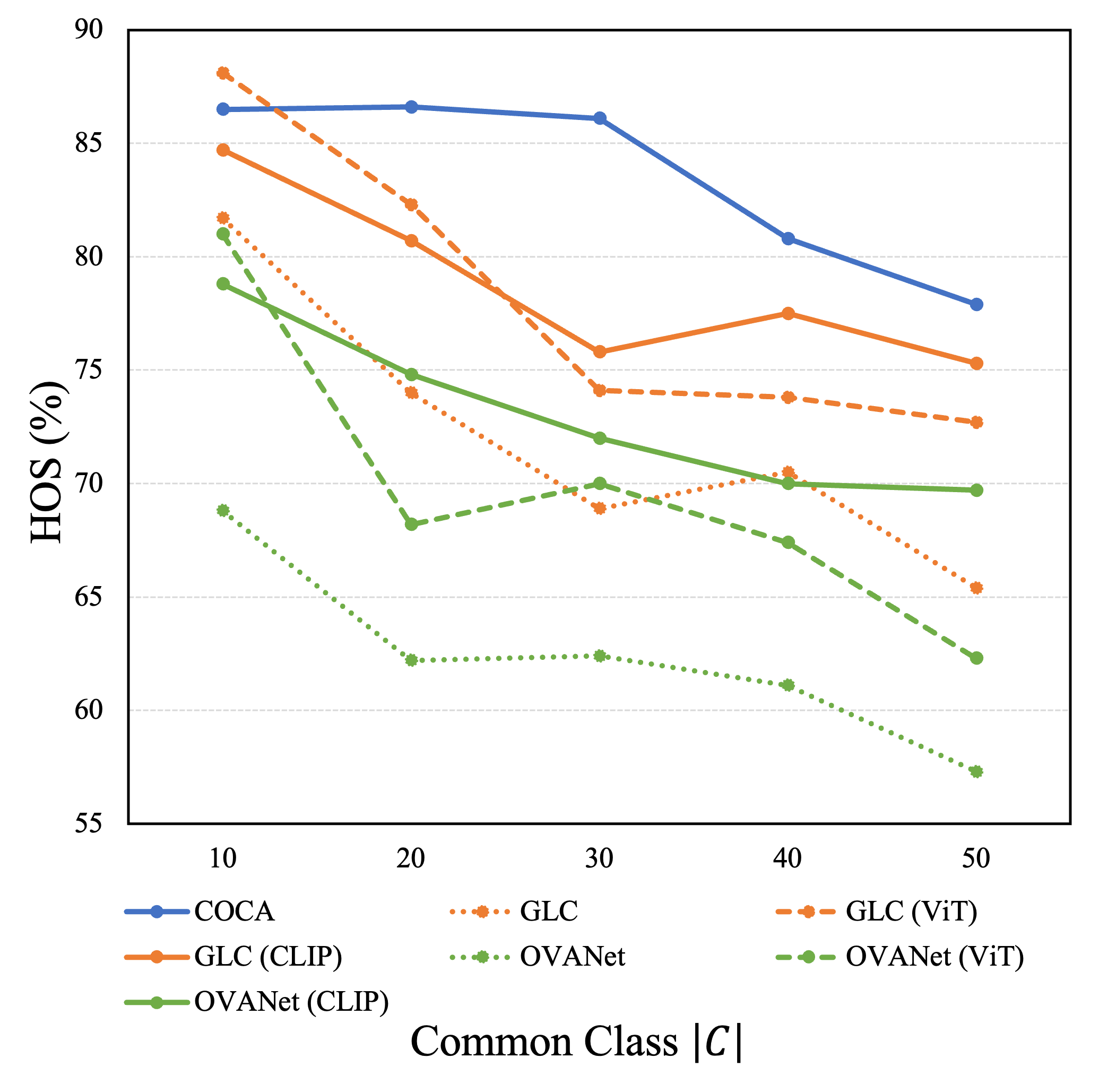}
         \caption{Cl$\rightarrow$Pr in OPDA}
         \label{Fig:Cl2Pr}
     \end{subfigure}
     \hfill
     \begin{subfigure}[b]{0.32\textwidth}
         \centering
         \includegraphics[width=\textwidth]{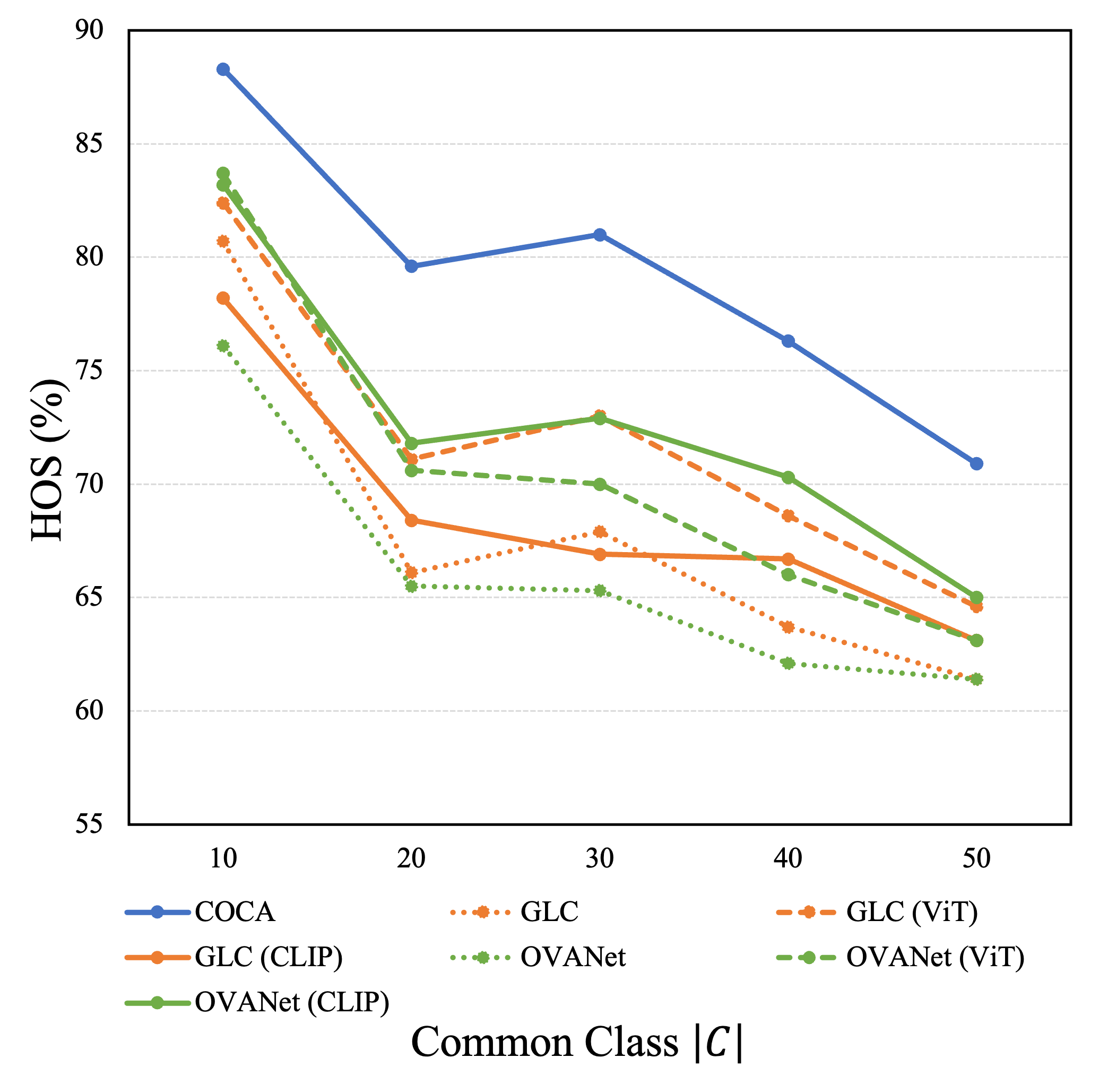}
         \caption{Rw$\rightarrow$Ar in OPDA}
         \label{Fig:Rw2Ar}
     \end{subfigure}
     \hfill
     \begin{subfigure}[b]{0.32\textwidth}
         \centering
         \includegraphics[width=\textwidth]{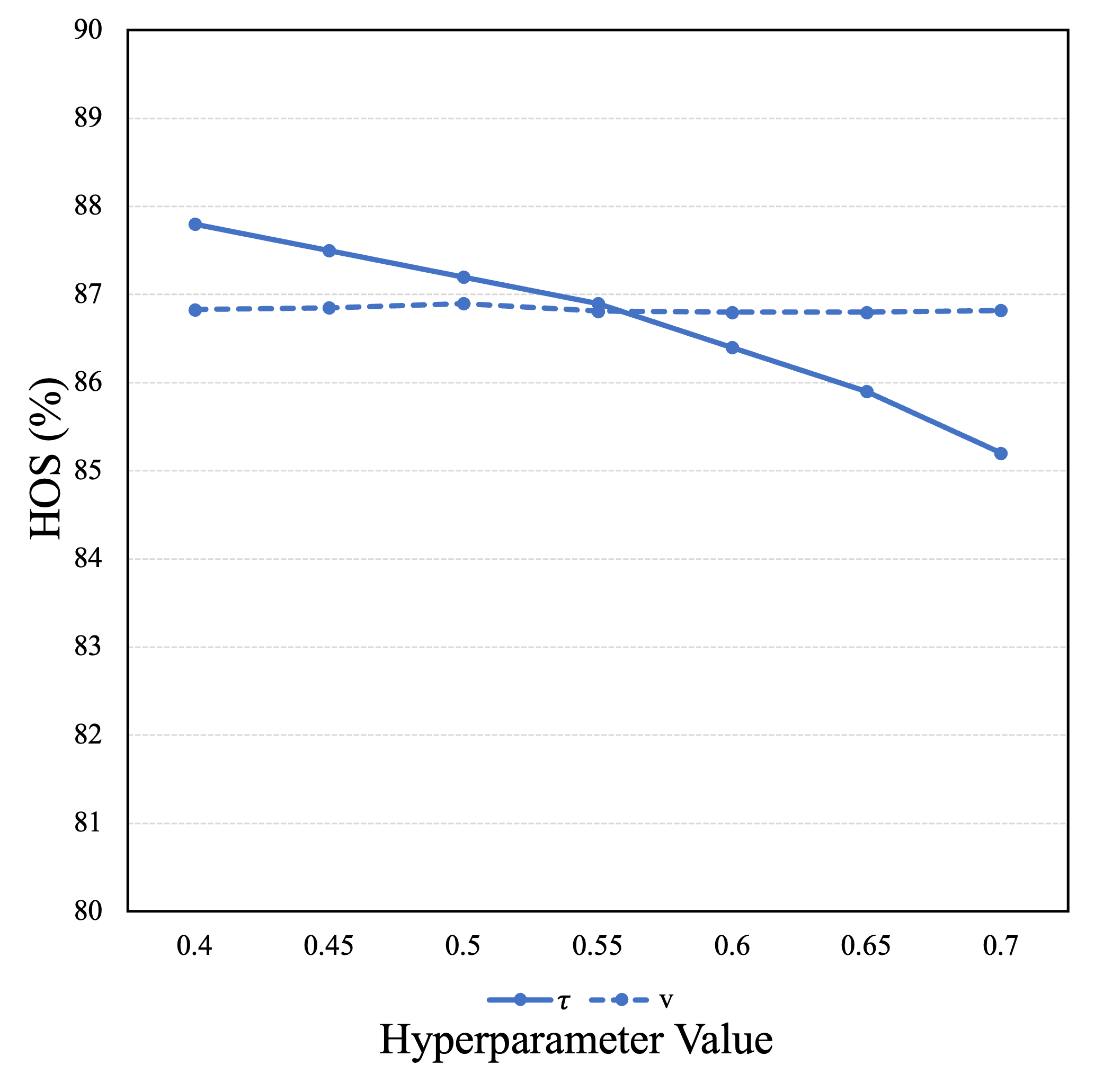}
         \caption{HOS w.r.t $\tau$ and $\mathsf{v}$}
         \label{Fig:hyperparameter}
     \end{subfigure}
     \caption{(a-b)HOS (\%) with respect to the number of common class $|C|$ on OfficeHome in OPDA. (c) HOS (\%) with respect to $\tau$ and $\mathsf{v}$ on OfficeHome in OPDA.}
\end{figure}

\textbf{Impact of Varying $|C|$.} We evaluate the robustness of COCA by contrasting it with other methods under varying numbers of common classes $|C|$ on OfficeHome in OPDA. \cref{Fig:Cl2Pr} and \cref{Fig:Rw2Ar} illustrate that COCA overall outperforms and demonstrates greater stability than preceding models.

\begin{figure}[t]
\centering
     \begin{subfigure}[b]{0.32\textwidth}
         \centering
         \includegraphics[width=\textwidth]{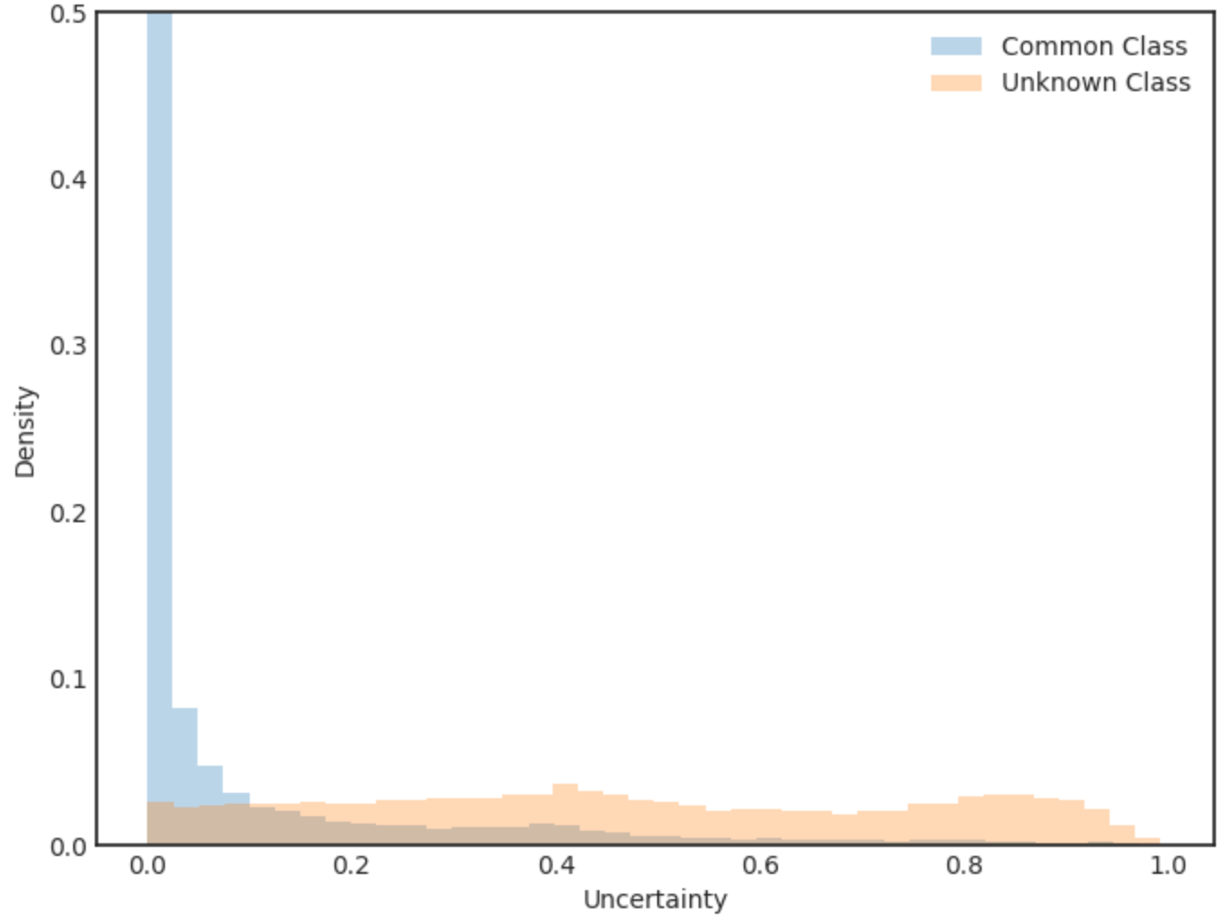}
         \caption{Source Model (16-shot)}
     \end{subfigure}
     \hfill
     \begin{subfigure}[b]{0.32\textwidth}
         \centering
         \includegraphics[width=\textwidth]{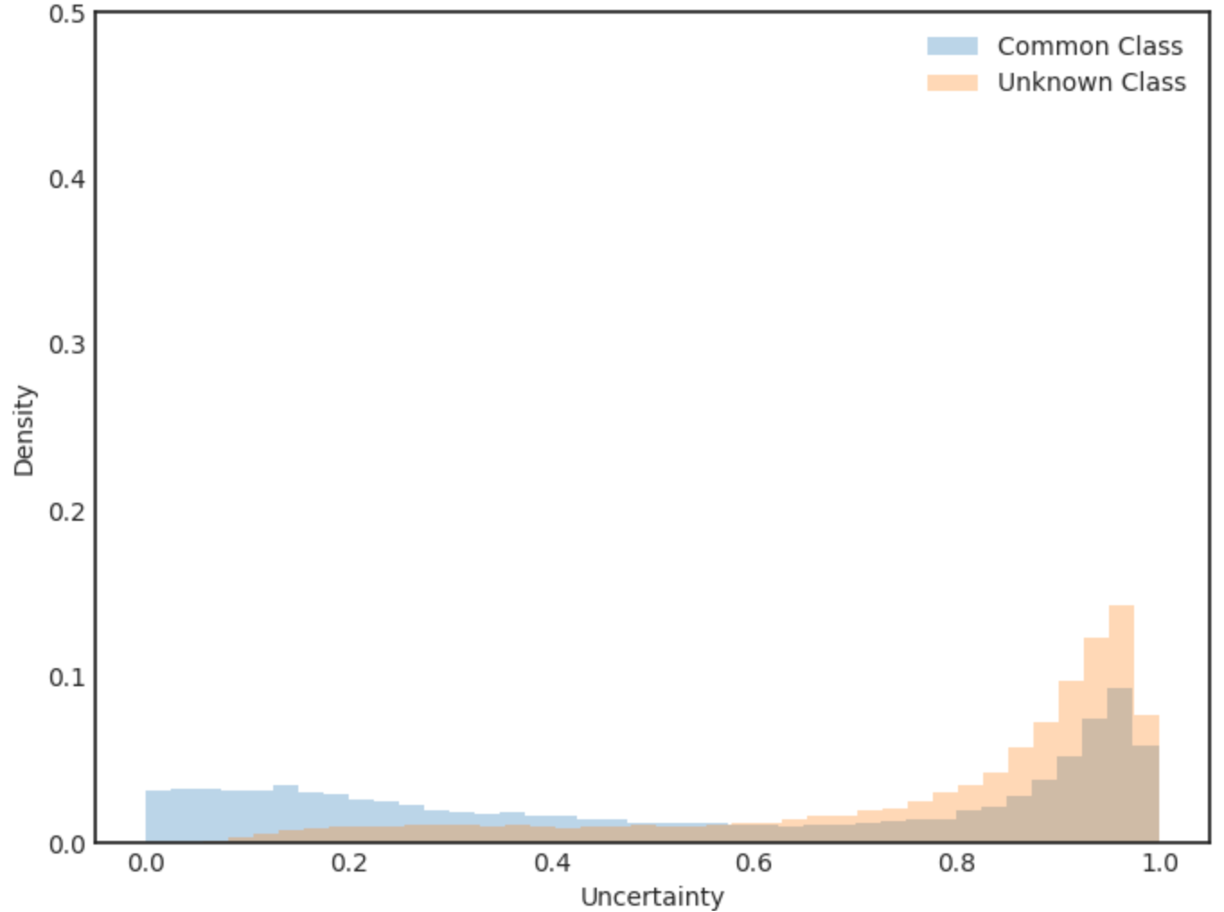}
         \caption{+ COCA-w-$p^c$}
     \end{subfigure}
     \hfill
     \begin{subfigure}[b]{0.32\textwidth}
         \centering
         \includegraphics[width=\textwidth]{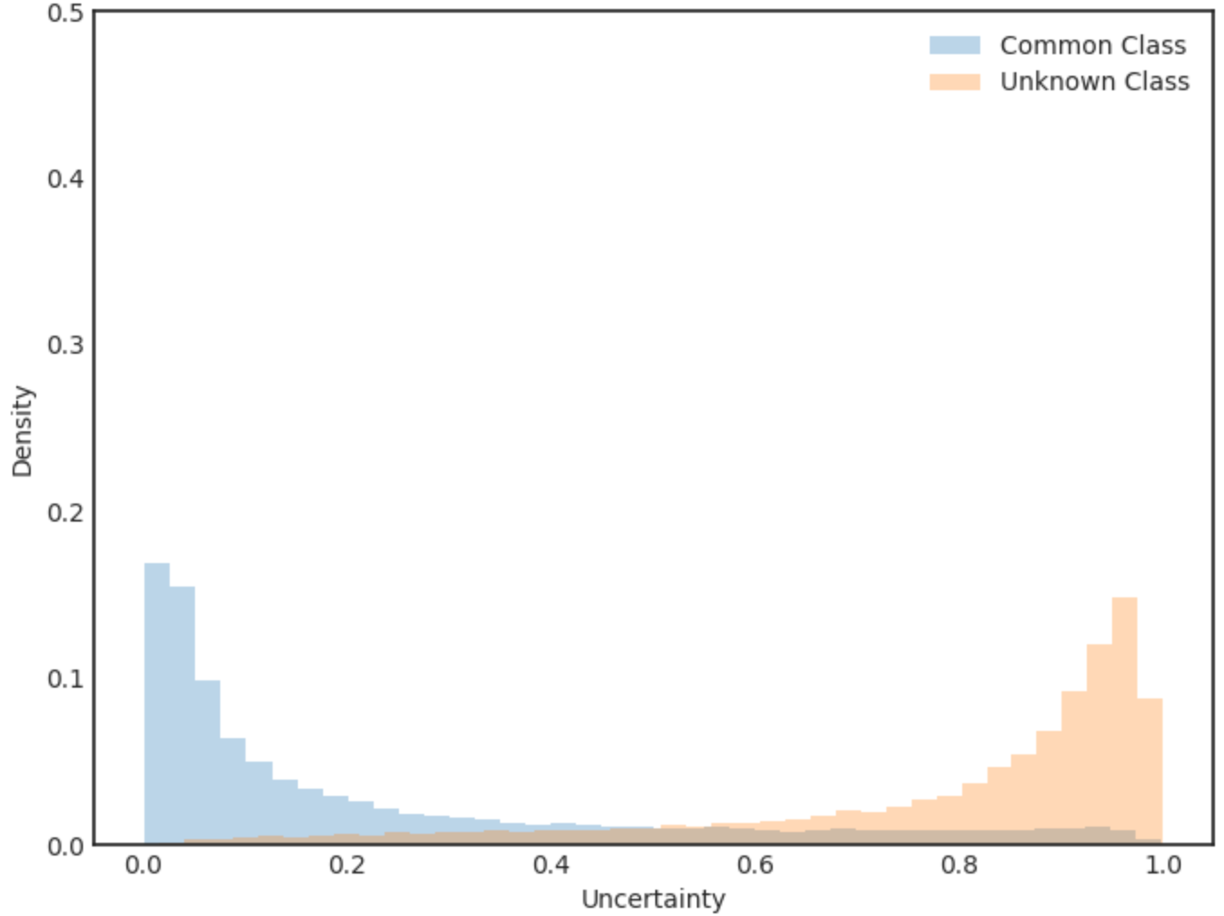}
         \caption{+ COCA}
     \end{subfigure}
     \caption{Uncertainty distribution of the source model \cite{cross-modal_adaptation}, source model + COCA-w-$p^c$, and source model + COCA for common and unknown class images on VisDA-2017 in OSDA.}
     \label{fig:uncertainty}
\end{figure}

\textbf{Ablation Study.}
We conducted comprehensive ablation studies on the three datasets to assess the effectiveness of distinct components within our method.
The results are summarized in \cref{tab:ablation_study}, where $OS=\frac{|C^s|}{|C^s|+1}\times OS^* + \frac{|C^s|}{|C^s|+1} \times UNK$ indicates the average accuracy on different classes. 
Compared to \textbf{COCA-w-$p^c$}, \textbf{COCA} shows 3.1\% improvement in HOS for OPDA on OfficeHome, 3.4\% on VisDA, and 1.3\% on DomainNet. It indicates that textual prototypes $z^\text{img}_c$ are more appropriate than image prototypes $p^c$ for positive prototypes due to $R_\text{IB}(Z^\text{text})>R_\text{IB}(V^\text{img})$, as discussed in our paper (\cref{eq:ib_comparison}). 
\textcolor{black}{\textbf{COCA-w/o-$h_\theta$} represents the combination of the ACTP module and the zero-shot CLIP without the linear classifier $h_\theta$.}
The HOS results of \textbf{COCA-w/o-$h_\theta$} highlight the potential of integrating image and text encoders within VLMs. This integration enables the precise separation of common and unknown class samples.
However, a considerable performance gap remains when compared to the full \textbf{COCA} method. The \textbf{COCA} method demonstrates significant improvements, achieving a 3.2\% increase in HOS for OPDA on OfficeHome, a 5.5\% improvement on VisDA, and a remarkable 9.3\% enhancement on DomainNet.
The result gaps of \textbf{COCA} and \textbf{COCA-w/o-$h_\theta$} show our method's effectiveness. The innovative paradigm we propose in our paper (\cref{Fig:Pipeline_comparison}), emphasizing classifier optimization rather than the image encoder optimization seen in previous UniDA/SF-UniDA methods, presents a more fitting approach based on VLMs to tackle SF-UniDA challenges as we discussed in our paper (\cref{Dis:reason_adapt_decision_boundary}).
\textbf{COCA-w/o-MIECI} indicates the removal of the MIECI module. A comparative analysis of results between \textbf{COCA-w/o-MIECI} and \textbf{COCA} reveals that the MIECI module plays a crucial role in promoting the learning of context relations within target images. This results in an increase in mutual information $I(Z^\text{img}, Y; h_\theta)$. As discussed in our paper (\cref{Dis:reason_mieci}), this improvement directly contributes to enhanced model performance, specifically in terms of accuracy in classifying common class samples. To visually assess the separation between the common and unknown classes on VisDA-2017 in OSDA, we present the uncertainty density distribution in \cref{fig:uncertainty}. The level of uncertainty indicates the extent to which the model regards the input image as belonging to an unknown class. The results demonstrate that while the source model performs well in classifying common classes, it struggles with the separation of unknown classes. In contrast, \textbf{COCA-w-$p^c$} exhibits imprecise recognition of common classes. Notably, \textbf{COCA} achieves a better balance between common class classification and unknown class identification, highlighting the superiority of textual prototypes.

\begin{table}[t]
  \centering
  \caption{HOS and OS (\%) of variants of COCA in \textbf{OPDA}.}
\label{tab:ablation_study}
  \resizebox{0.5\textwidth}{!}{
\begin{tabular}{l|cccccc}
\toprule
\cmidrule{2-7}
 & \multicolumn{2}{c}{OfficeHome}  & \multicolumn{2}{c}{VisDA-2017}  & \multicolumn{2}{c}{DomainNet}\\ 
\cmidrule{2-7}
 & OS & HOS & OS & HOS & OS & HOS\\
\midrule
COCA-w/o-$h_\theta$ & 88.8 & 83.7 & 83.0 & 77.7 & 50.3  & 63.8  \\
COCA-w/o-MIECI & 89.0 & 86.6 & 83.6  & 82.2 & 65.7 & 72.9 \\
COCA-w-$p^c$ & 81.0 & 83.8 & 74.7 & 79.8 & 66.2 & 71.8  \\
COCA & \textbf{90.2} & \textbf{86.9} & \textbf{85.2} & \textbf{83.2} & \textbf{66.4} & \textbf{73.1} \\
\bottomrule
\end{tabular}
}
\end{table}

\begin{table}[t]
\centering
\caption{HOS (\%) with respect to prompts and $K$ selecting methods in \textbf{OPDA}.}
 \begin{subtable}[t]{0.49\textwidth}
    \centering
    \subcaption{HOS (\%) comparison of various prompts.}
    \label{tab:HOS_comparison_various_prompts}
    \resizebox{\textwidth}{!}{
    \begin{tabular}{c|c|c|c}
    \toprule
    Prompt & OfficeHome & VisDA-2017 & DomainNet \\
    \midrule
    \texttt{a photo of a \{CLS\}} & 86.9 & 83.2 & 73.1  \\
    \texttt{a photo of some \{CLS\}} & 87.4 & 82.6 & 72.9 \\
    \texttt{a picture of a \{CLS\}} & 88.3 & 82.1 & 72.1 \\
    \texttt{a painting of a \{CLS\}} & 86.5 & 82.2 & 73.0 \\
    \texttt{this is a photo of a \{CLS\}} & 87.4 & 84.0 & 72.1 \\
    \texttt{this is a  \{CLS\} photo} & 86.6 & 83.3 & 72.5 \\
    \bottomrule
    \end{tabular}
    }
    \end{subtable}
\hfill
 \begin{subtable}[t]{0.49\textwidth}
    \centering
    \subcaption{HOS (\%) comparison of various $K$ selecting methods.}
    \label{tab:HOS_comparison_various_k_metric}
    \resizebox{\textwidth}{!}{
    \begin{tabular}{c|c|c|c}
    \toprule
    Method & OfficeHome & VisDA-2017 & DomainNet \\
    \midrule
    Calinski-Harabasz \cite{dendrite} & 86.9 & 82.7 & 72.8  \\
    Davies-Bouldi \cite{A_Cluster_Separation_Measure} & 86.9 & 83.2 & 73.0 \\
    Silhouette \cite{Silhouettes} & 86.9 & 83.2 & 73.1   \\
    \bottomrule
    \end{tabular}
    }
    \end{subtable}
\end{table}

\textbf{Hyperparameter Sensitivity.} \cref{Fig:hyperparameter} demonstrates the sensitivity to the hyperparameter $\tau$ and mask ratio $\mathsf{v}$ in OPDA on OfficeHome. The source model \cite{cross-modal_adaptation} + COCA is stable across a range of values for both $\tau$ and $\mathsf{v}$. 
The comparative experiments of prompts are shown in \cref{tab:HOS_comparison_various_prompts}, and our method exhibits stable performance across a variety of prompts.
We substitute the Silhouette method \cite{Silhouettes} with alternative methods, including the Calinski-Harabasz method \cite{dendrite} and the Davies-Bouldin method \cite{A_Cluster_Separation_Measure}, to ascertain the optimal $K$ value for the K-means clustering. This adjustment aims to evaluate COCA's generalization capabilities, with the results presented in \cref{tab:HOS_comparison_various_k_metric}. Comparing the results of Silhouettes, Calinski-Harabasz, and Davies-Bouldin methods, we deduce that COCA exhibits good generalization capabilities. This conclusion arises from the stable performance of COCA in OPDA across various methods used to determine the optimal $K$ value for the K-means clustering. The optimal $K$ value at the target domain adaptation phase selected by various methods \cite{dendrite,A_Cluster_Separation_Measure,Silhouettes} in OPDA is presented in \cref{tab:detailed_K_value_in_opda}.

\begin{table*}[t]
    \centering
    \caption{Optimal $K\in[1/3|C^s|,1/2|C^s|,|C^s|,2|C^s|,3|C^s|]$ selected by various methods in \textbf{OPDA}.}
    \label{tab:detailed_K_value_in_opda}
    \resizebox{0.9\textwidth}{!}{
    \begin{tabular}{c|*{4}{w{c}{1.5cm}}|c|*{3}{w{c}{1.8cm}}}
    \toprule
    \multirow{2}{*}{Method} & \multicolumn{4}{c|}{OfficeHome ($|C^s|=15,|C^t|=50$)} & \multirowcell{2}{VisDA-2017\\($|C^s|=9,|C^t|=9$)}& \multicolumn{3}{c}{DomainNet ($|C^s|=200,|C^t|=195$)} \\
    \cmidrule{2-5} \cmidrule{7-9}
    & $\rightarrow$Ar & $\rightarrow$Cl & $\rightarrow$Pr & $\rightarrow$Rw & & $\rightarrow$P & $\rightarrow$R & $\rightarrow$S\\
    \midrule
    Calinski-Harabasz \cite{dendrite} & 45 & 45 & 45 & 45 & 27 & 600 & 600 & 600 \\
    Davies-Bouldi \cite{A_Cluster_Separation_Measure} & 45 & 30 & 45 & 45 & 9 & 200 & 200 & 200 \\
    Silhouette \cite{Silhouettes} & 45 & 45 & 45 & 45 & 9 & 200 & 400 & 400 \\
    \bottomrule
    \end{tabular}
    }
\end{table*}

\textbf{Boxplots.} An illustration of boxplots with 5 different random seeds in Fig.\ref{fig:hos_boxplot} demonstrates that COCA achieves more accurate performance in separating common and unknown classes than existing methods.

%% file: eccv24/appendix/implementation_details.tex
\subsection{Source Model Details}
\begin{table}[t]
\centering
\caption{Batch size for source model training.}
\resizebox{0.25\textwidth}{!}{
\begin{tabular}{l|c}
\toprule
 & batch size \\ 
\midrule
$8 \leq 2|C^s| < 16$ & 8 \\
$16 \leq 2|C^s| < 32$ & 16 \\
$32 \leq 2|C^s| < 64$ & 32 \\
$64 \leq 2|C^s|$ & 64 \\
\bottomrule
\end{tabular}
}
\label{tab:batch_size_source_model}
\end{table}

\begin{figure*}[t]
\centering
     \includegraphics[width=0.7\textwidth]{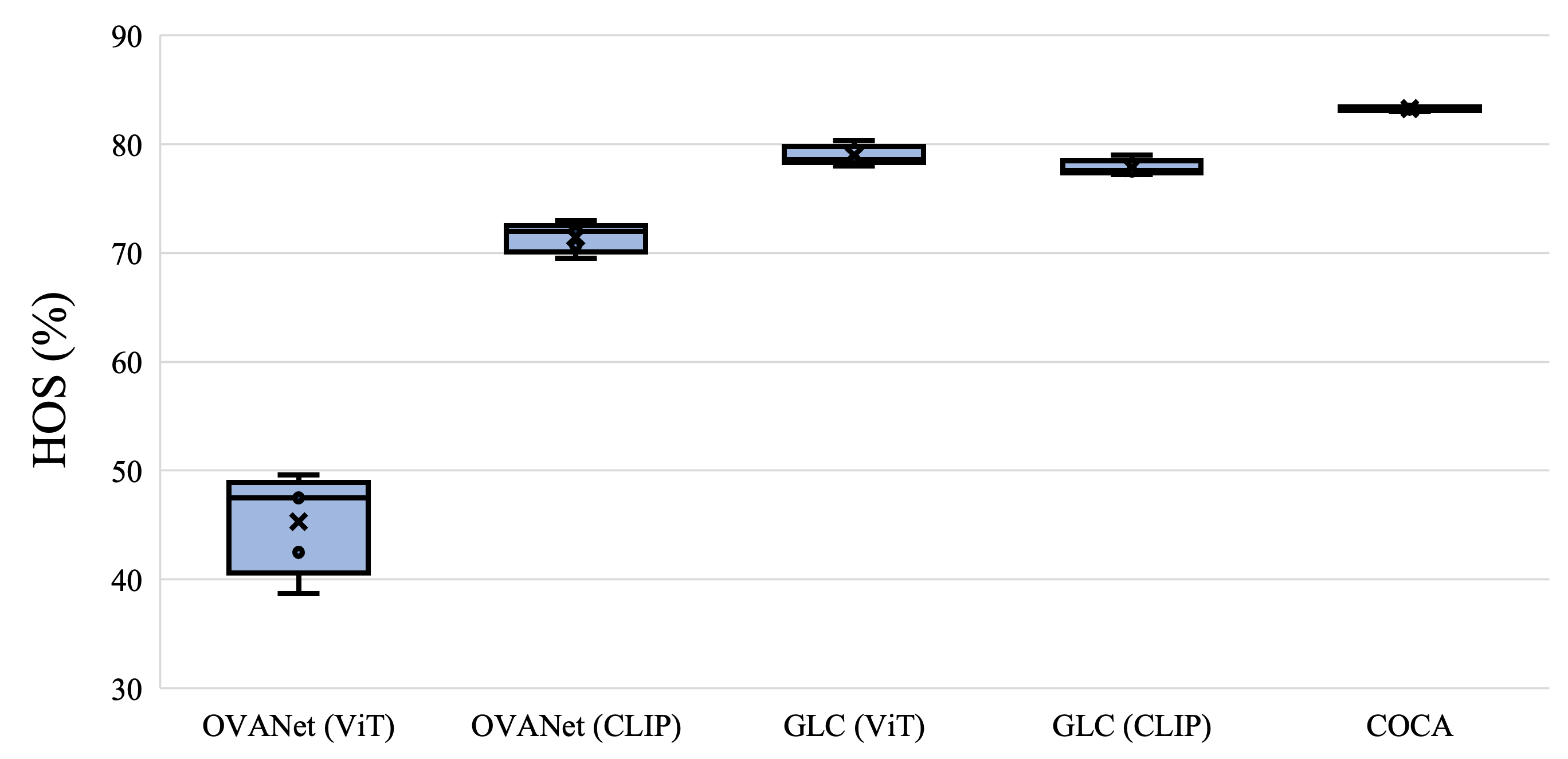}
     \caption{HOS rate of OVANet (ViT), OVANet (CLIP), GLC (ViT), GLC (CLIP), and COCA on VisDA-2017 in OPDA. Each boxplot ranges from the upper and lower quartiles with the median as the horizontal line and whiskers extend to 1.5 times the interquartile range.}
     \label{fig:hos_boxplot}
\end{figure*}

The source model is linear probe CLIP \cite{CLIP}, CLIP-Adapter \cite{gao2021clipadapter}, or cross-modal linear probing \cite{cross-modal_adaptation} based on CLIP(ViT-B/16). At the source model training phase, we freeze the image and text encoders and optimize the classifier. The classifier of linear probe CLIP \cite{CLIP} or cross-model linear probing \cite{cross-modal_adaptation} is the single linear layer. The classifier of CLIP-Adapter \cite{gao2021clipadapter} is the adapter module. The basic settings are as follows: (1) Performing a learning rate warmup with 50 iterations, during which the learning rate goes up linearly from 0.00001 to the initial value. (2) Performing a cosine annealing learning rate scheduling over the course of 12800 iterations. (3) Employing early stopping based on the few-shot validation set performance evaluated every 100 iterations.

We have established the batch size for source model training as outlined in \cref{tab:batch_size_source_model}. We configure the weight decay to 0.01 for all benchmarks and the initial learning rate to 0.001 for OfficeHome \cite{OfficeHome} and VisDA-2017 \cite{VisDa}, and 0.0001 for DomainNet \cite{DonmaiNet}. The optimizer of the source model is AdamW \cite{AdamW}. Given that the cross-modal linear probing model \cite{cross-modal_adaptation} necessitates the inclusion of varied class names within a mini-batch for training, and that the input is comprised of a 50-50 split between images and text. For the CLIP-Adapter model \cite{gao2021clipadapter}, we follow the same 2-layer MLP architecture with the given residual ratio of 0.2.

\textbf{Image Loss.} Given a image feature $z^{\text{img,s}}_i$ for the source image $x^s_i$ and the corresponding ground truth label $y^s_i$, the image loss $\mathcal{L}^{\text{img}}_s$ for the source model training is
$\mathcal{L}^{\text{img}}_s=-\frac{1}{N^s}\sum_{\substack{i=1}}^{\substack{N^s}}y^s_i\log(\sigma(h_\theta(z^{\text{img,s}}_i)))$,
where $N^s$ is the number of source samples.

\textbf{Text Loss.} Given a text feature $z^\text{text}_c$ converted from text template \texttt{a photo of a \{CLS\}}, the corresponding ground truth label $y_c$, the text loss $\mathcal{L}^{\text{text}}_s$ for the cross-modal linear probing model is
$\mathcal{L}^{\text{text}}_s=-\frac{1}{|C^s|}\sum_{\substack{c=1}}^{\substack{|C^s|}}y_c\log\left(\sigma\left(h_\theta\left(z^\text{text}_c\right)\right)\right)$.

\textbf{Overall Loss.} \textcolor{black}{For linear probe CLIP and CLIP-Adapter models, the overall loss $\mathcal{L}_s$ for source model training is
$\mathcal{L}_s=\mathcal{L}^{\text{img}}_s$.
For the cross-modal linear probing model, the overall loss $\mathcal{L}_s$ for source model training is $\mathcal{L}_s=\mathcal{L}^{\text{img}}_s+\mathcal{L}^{\text{text}}_s$.}

\subsection{Silhouette Score}
For a image feature $z^\text{img}_i \in \mathcal{C_I}$, where $\mathcal{C_I}$ is one of the $K$ clusters, computing the mean distance between $z^\text{img}_i$ and other image features $z^\text{img}_j$ within the same cluster as follows:
\begin{equation}
    a(z^\text{img}_i)=\frac{1}{|\mathcal{C_I}|-1}\displaystyle\sum_{\substack{z^\text{img}_j\in \mathcal{C_I}, i\neq j}}d(z^\text{img}_i,z^\text{img}_j),
\end{equation}
where $|\mathcal{C_I}|$ denotes the number of image features belonging to cluster $\mathcal{C_I}$, and $d(z^\text{img}_i,z^\text{img}_j)$ is the distance between $z^\text{img}_i$ and $z^\text{img}_j$ within the cluster $\mathcal{C_I}$.
\begin{equation}
b(z^\text{img}_i)=\displaystyle\min_{\substack{\mathcal{J}\neq \mathcal{I}}}\frac{1}{|\mathcal{C_J}|}\displaystyle\sum_{\substack{z^\text{img}_j\in \mathcal{C_J}}}d(z^\text{img}_i,z^\text{img}_j)
\end{equation}
is the distance between $z^\text{img}_i$ and the "neighboring cluster" of $z^\text{img}_i$. The mean distance from $z^\text{img}_i$ to all image features $z^\text{img}_j$ in $\mathcal{C_J}$ is calculated as the dissimilarity of $z^\text{img}_i$ to another cluster $\mathcal{C_J}$, where $\mathcal{C_J}\neq \mathcal{C_I}$. 
The Silhouette score $s(z^\text{img}_i)$ is defined as:
\begin{equation}
s(z^\text{img}_i)=\frac{b(z^\text{img}_i)-a(z^\text{img}_i)}{\max\{a(z^\text{img}_i),b(z^\text{img}_i)\}}.
\label{eq:silhouette_value}
\end{equation}
High Silhouette scores for the majority of image features suggest that the K-means hyperparameter $K$ value is well-chosen, indicating that image features within the same cluster are closely grouped and well-separated from those in other clusters.

\subsection{Pseudo Code}
The training procedure of the proposed method is summarized in \cref{algo:proposed_method}.

\begin{algorithm}[t]
\caption{Traning Producedure of the Proposed Method}\label{algo:proposed_method}
\begin{algorithmic}[1]
\Require Target domain dataset $D^t=\left\{x_i\right\}_{i=1}^N$, prompts \verb+a photo of a {CLS}+, image encoder $G^\text{img}$, text encoder $G^\text{text}$, source model's classifier $h_\theta^s$, $K$ candidate list $[1/3|C^s|, 1/2|C^s|, |C^s|, 2|C^s|, 3|C^s|]$, and other necessary hyperparameters
\Ensure Target domain classifier $h_\theta$
\State Freeze $G^\text{img}$ and $G^\text{text}$
\State Input $D^t$ to $G^\text{img}$ to generate target-image features $Z^\text{img}=\left\{z_i^\text{img}\right\}_{i=1}^N$
\State $bestK \leftarrow 0$, $maxScore \leftarrow 0$
\For{$candidateK \in [1/3|C^s|, 1/2|C^s|, |C^s|, 2|C^s|, 3|C^s|]$}
    \State $K \leftarrow candidateK$ \Comment{K-means hyperparameter}
    \State Input $Z^\text{img}$ to K-means to cluster all target image features
    \State Calculate target image features' Silhouette score $s\left(z^\text{img}\right)$ \Comment{\cref{eq:silhouette_value}}
    \State Take an average score $\overline{s}=\frac{1}{N}\sum_{i=1}^N s\left(z^\text{img}_i\right)$
    \If{$\overline{s}>maxScore$}
        \State $bestK \leftarrow candidateK$, $maxScore \leftarrow \overline{s}$
    \EndIf
\EndFor
\State $h_\theta \leftarrow h_\theta^s$, $h_\gamma \leftarrow h_\theta^s$ \Comment{Initialize $h_\theta$ and $h_\gamma$}
\State Input prompts to $G^\text{text}$ to generate text features $Z^\text{text}=\left\{z_c^\text{text}\right\}_{c=1}^{|C^s|}$
\State $K \leftarrow bestK$ \Comment{K-means hyperparameter}
\State Input $Z^\text{img}$ to K-means to generate image prototypes $\{v_k^\text{img}\}_{k=1}^K$ \Comment{\cref{eq:generate_image_prototype}}
\State Determine negative image prototypes $\{n^c_k\}_{k=1}^{K-1}$ for known class $c$ \Comment{\cref{eq:positive_negative_image_prototypes}}
\State Generate a pseudo label $\hat{y}_i$ for each target image $x_i$ \Comment{\cref{eq:generate_pseudo_label}}
\For{$epoch = 1$ \textbf{to} $maxEpoch$}
    \State Calculate the image cross-entropy loss $\mathcal{L}^{\text{img}}$ \Comment{\cref{eq:image_ce_loss}}
    \State Calculate the text cross-entropy loss $\mathcal{L}^{\text{text}}$ \Comment{\cref{eq:text_ce_loss}}
    \State Generate patch mask $\mathsf{M}$ and masked target image $x^\mathsf{M}_i$ \Comment{\cref{eq:generate_patch_mask} and \cref{eq:generate_masked_target_image}}
    \State Calculate the mask loss $\mathcal{L}^\text{mask}$ \Comment{\cref{eq:cal_mask_loss}}
    \State $\theta \leftarrow \theta - \nabla_\theta (\mathcal{L}^{\text{img}}+\mathcal{L}^{\text{text}}+\mathcal{L}^\text{mask}) $ \Comment{Update $h_\theta$}
    \State $\gamma \leftarrow \alpha\gamma + (1-\alpha)\theta$ \Comment{Update teacher classifier $h_\gamma$}
\EndFor
\end{algorithmic}
\end{algorithm}

\subsection{Baseline Details}
We have reproduced several open-source UniDA/SF-UniDA models, and the details of the parameters are provided below:

\textbf{DCC.} We use CLIP(ViT-B/16) \cite{CLIP} as the backbone. The classifier is made up of two FC layers. We use Nesterov momentum SGD to optimize the model, which has a momentum of $0.9$ and a weight decay of $5\text{e-}4$. The learning rate decreases by a factor of  $(1+\alpha \frac{i}{N})^{-\beta}$, where $i$ and $N$ represent current and global iteration, respectively, and we set $\alpha = 10$ and $\beta = 0.75$. We use a batch size of 36, and the initial learning rate is set as $1\text{e-}4$ for Office-31, and $1\text{e-}3$ for OfficeHome and DomainNet. We use the settings detailed in the original paper \cite{DCC}. PyTorch \cite{Pytorch} is used for implementation.

\textbf{OVANet.} For OVANet \cite{OVANet} with ViT-B/16 \cite{ViT} and CLIP(ViT-B/16) backbones, we adopt the hyperparameter settings outlined in the original paper \cite{OVANet}. Specifically, we utilize inverse learning rate decay scheduling for the learning rate schedule and assign a weight of $\lambda=0.1$ for the entropy minimization loss across all benchmarks. The batch size is fixed at 36, with the initial learning rate set to 0.01 for the classification layer and 0.001 for the backbone layers. PyTorch \cite{Pytorch} is used for the implementation.

\textbf{GLC.} For GLC \cite{GLC} with ViT-B/16 and CLIP(ViT-B/16), we employ the SGD optimizer with a momentum of 0.9 at the target model adaptation phase. The initial learning rate is set to 0.001 for OfficeHome and 0.0001 for both VisDA-2017 and DomainNet. The hyperparameter $\rho$ is fixed at 0.75 and $|L|$ at 4 across all datasets, while $\eta$ is set to 0.3 for VisDA and 1.5 for OfficeHome and DomainNet. All these hyperparameters correspond to the settings detailed in the original paper \cite{GLC}. PyTorch is used for the implementation.

%% file: eccv24/appendix/discussion.tex
\subsection{K-means Clustering Invocations}
\begin{table}[t]
\caption{The number of calls of K-means clustering in OPDA. 100 clustering iterations per call.}
\label{tab:cluster_times}
\centering
\resizebox{0.5\textwidth}{!}{
\begin{tabular}{l|c|c|c}
\toprule
 & OfficeHome  & VisDA-2017  & DomainNet  \\ 
\midrule
GLC & 15 $\times maxEpoch$ & 9 $\times maxEpoch$ & 200 $\times maxEpoch$ \\
Ours & 1 & 1  & 1 \\
\bottomrule
\end{tabular}
}
\end{table}

In this subsection, we will discuss the frequency of K-means clustering invocations per epoch in OPDA with that of GLC \cite{GLC}. 

As shown in \cref{tab:cluster_times}, compared to GLC \cite{GLC}, our method significantly reduces the times of K-means clustering. Our method merely needs to cluster all image features once, and then it can identify the negative prototypes for all known classes. In contrast, the GLC model must apply the K-means cluster $|C^s|$ times per epoch to locate negative prototypes for all known classes. This suggests that our methods can save significant time on large-scale datasets, particularly when $|C^s|$ is large.

GLC employs the Top-K method to obtain positive image features for a known class $c$. The hyperparameter of Top-K is represented as $K'$ to differentiate it from the K-means hyperparameter $K$. After implementing Top-K for each known class, GLC obtains a positive image feature set $\{z^\text{img, pos}_{c,i}\}_{i=1}^{K'}$, where $z^\text{img, pos}_{c,i}$ symbolizes the positive image feature for a known class $c$ and a negative image feature set $\{z^\text{img, neg}_{c,j}\}_{j=1}^{N-K'}=\{z^\text{img}_l\}_{l=1}^{N}/\{z^\text{img, pos}_{c,i}\}_{i=1}^{K'}$, where $z^\text{img, neg}_{c,j}$ signifies the negative image feature and $\{z^\text{img}_l\}_{l=1}^{N}$ represents the target image feature set, with $N$ being the number of target samples. As the positive image feature set varies for each known class $c$, so too does the negative image feature set for each respective class. Thus, GLC needs to invoke the K-means clustering $|C^s|$ times to obtain the negative image prototype sets $\{\{n^c_m\}^{K-1}_{m=1}\}^{|C^s|}_{c=1}$ for all known classes. For instance, consider six image features $\{z^\text{img}_l\}_{l=1}^6$, two known classes $\{c_1,c_2\}$ and the unknown class \texttt{unknown}, where $\{z^\text{img}_1, z^\text{img}_2\}$ belong to $c_1$, $\{z^\text{img}_3, z^\text{img}_4\}$ to $c_2$, and $\{z^\text{img}_5, z^\text{img}_6\}$ to \texttt{unknown}. GLC uses Top-K $(K'=2)$ to select the positive image features $\{z^\text{img,pos}_{c_1,i}\}_{i=1}^2=\{z^\text{img}_1,z^\text{img}_2\}$ for the class $c_1$ and $\{z^\text{img,pos}_{c_2,i}\}_{i=1}^2=\{z^\text{img}_3,z^\text{img}_4\}$ for the class $c_2$; the negative image features for $c_1$ are $\{z^\text{img,neg}_{c_1,j}\}_{j=1}^4=\{z^\text{img}_3,z^\text{img}_4,z^\text{img}_5,z^\text{img}_6\}$ and for $c_2$ are $\{z^\text{img,neg}_{c_2,j}\}_{j=1}^4=\{z^\text{img}_1,z^\text{img}_2,z^\text{img}_5,z^\text{img}_6\}$. Given that the negative image feature sets $\{\{z^\text{img, neg}_{c,j}\}_{j=1}^{4}\}_{c=c_1}^{c_2}$ varies for the known classes $c_1,c_2$, GLC requires to invoke K-means clustering $|C^s|=2$ times to generate the negative image prototype sets $\{n^{c_1}_m\}_{m=1}^{K-1}=\text{K-means}\left(\{z^\text{img, neg}_{c_1,j}\}_{j=1}^{4}\right)$ and $\{n^{c_2}_m\}_{m=1}^{K-1}=\text{K-means}\left(\{z^\text{img, neg}_{c_2,j}\}_{j=1}^{4}\right)$ for all known classes. Furthermore, in GLC, since both the image encoder and the bottleneck layer—situated between the image encoder and the classifier for local census clustering—require updates at each epoch, the K-means clustering must be invoked at each epoch.

On the other hand, in our approach, the target image feature set $\{z^\text{img}_l\}_{l=1}^{N}$ remains constant. We first apply K-means to $\{z^\text{img}_l\}_{l=1}^{N}$ to derive all image prototypes $\{v_i\}_{i=1}^K=\text{K-means}\left(\{z^\text{img}_l\}_{l=1}^{N}\right)$. Subsequently, we perform matrix multiplication between the text feature $z^\text{text}_c$ of known class $c$ and the image prototype set $\{v_i\}_{i=1}^K$ to identify positive and negative image prototypes. As matrix multiplication is considerably more efficient than K-means, our approach significantly reduces computational time in comparison to GLC. In our method, we only need to invoke the K-means clustering at the first epoch since the image and text encoders are frozen.

\subsection{Potential Societal Impact}
Our method can adapt a trained few-shot learner to unlabeled target datasets with uncertainty domain and category shifts by optimizing the classifier. In numerous instances where source datasets are unobtainable and the quantity of source samples is restricted, our approaches do not need to directly access source samples and substantially reduce the label cost of source samples. This might make technology more accessible to organizations and individuals with limited resources. However, one potential downside is the increased availability of the systems to those seeking to exploit them for unlawful purposes. While we report an enhanced performance in comparison to the current state-of-the-art methods, the results remain unsatisfactory in extreme scenarios of domain shift or category shift. Thus, our approach should not be deployed in critical applications or for making significant decisions without human supervision.

%% file: main.bbl
\begin{thebibliography}{10}
\providecommand{\url}[1]{\texttt{#1}}
\providecommand{\urlprefix}{URL }
\providecommand{\doi}[1]{https://doi.org/#1}

\bibitem{information_bottleneck}
Alemi, A.A., Fischer, I., Dillon, J.V., Murphy, K.: Deep variational information bottleneck. In: Int. Conf. Learn. Represent. (2017)

\bibitem{ROS}
Bucci, S., Loghmani, M.R., Tommasi, T.: On the effectiveness of image rotation for open set domain adaptation. In: Eur. Conf. Comput. Vis. pp. 422--438 (2020)

\bibitem{dendrite}
Caliński, T., Harabasz, J.: A dendrite method for cluster analysis. Communications in Statistics  \textbf{3}(1),  1--27 (1974)

\bibitem{PADA}
Cao, Z., Ma, L., Long, M., Wang, J.: Partial adversarial domain adaptation. In: Eur. Conf. Comput. Vis. pp. 135--150 (2018)

\bibitem{ETN}
Cao, Z., You, K., Long, M., Wang, J., Yang, Q.: Learning to transfer examples for partial domain adaptation. In: IEEE Conf. Comput. Vis. Pattern Recog. pp. 2985--2994 (2019)

\bibitem{UniOT}
Chang, W., Shi, Y., Tuan, H., Wang, J.: Unified optimal transport framework for universal domain adaptation. In: Adv. Neural Inform. Process. Syst. pp. 29512--29524 (2022)

\bibitem{GATE}
Chen, L., Lou, Y., He, J., Bai, T., Deng, M.: Geometric anchor correspondence mining with uncertainty modeling for universal domain adaptation. In: IEEE Conf. Comput. Vis. Pattern Recog. pp. 16134--16143 (2022)

\bibitem{A_Cluster_Separation_Measure}
Davies, D.L., Bouldin, D.W.: A cluster separation measure. IEEE Trans. Pattern Anal. Mach. Intell.  \textbf{1}(2),  224–227 (1979)

\bibitem{ImageNet}
Deng, J., Dong, W., Socher, R., Li, L.J., Li, K., Fei-Fei, L.: Imagenet: A large-scale hierarchical image database. In: IEEE Conf. Comput. Vis. Pattern Recog. pp. 248--255 (2009)

\bibitem{ViT}
Dosovitskiy, A., Beyer, L., Kolesnikov, A., Weissenborn, D., Zhai, X., Unterthiner, T., Dehghani, M., Minderer, M., Heigold, G., Gelly, S., Uszkoreit, J., Houlsby, N.: An image is worth 16x16 words: Transformers for image recognition at scale. In: Int. Conf. Learn. Represent. (2021)

\bibitem{CMU}
Fu, B., Cao, Z., Long, M., Wang, J.: Learning to detect open classes for universal domain adaptation. In: Eur. Conf. Comput. Vis. pp. 567--583 (2020)

\bibitem{DANN}
Ganin, Y., Ustinova, E., Ajakan, H., Germain, P., Larochelle, H., Laviolette, F., Marchand, M., Lempitsky, V.S.: Domain-adversarial training of neural networks. J. Mach. Learn. Res.  \textbf{17},  59:1--59:35 (2016)

\bibitem{gao2021clipadapter}
Gao, P., Geng, S., Zhang, R., Ma, T., Fang, R., Zhang, Y., Li, H., Qiao, Y.: Clip-adapter: Better vision-language models with feature adapters (2021)

\bibitem{AR}
Gu, X., Yu, X., yang, y., Sun, J., Xu, Z.: Adversarial reweighting for partial domain adaptation. In: Adv. Neural Inform. Process. Syst. pp. 14860--14872 (2021)

\bibitem{ResNet}
He, K., Zhang, X., Ren, S., Sun, J.: Deep residual learning for image recognition. In: IEEE Conf. Comput. Vis. Pattern Recog. pp. 770--778 (2016)

\bibitem{MIC}
Hoyer, L., Dai, D., Wang, H., Van~Gool, L.: {MIC}: Masked image consistency for context-enhanced domain adaptation. In: IEEE Conf. Comput. Vis. Pattern Recog. (2023)

\bibitem{UADAL}
Jang, J., Na, B., Shin, D.H., Ji, M., Song, K., Moon, I.c.: Unknown-aware domain adversarial learning for open-set domain adaptation. In: Adv. Neural Inform. Process. Syst. pp. 16755--16767 (2022)

\bibitem{SPA}
Kundu, J.N., Bhambri, S., Kulkarni, A.R., Sarkar, H., Jampani, V., R, V.B.: Subsidiary prototype alignment for universal domain adaptation. In: Adv. Neural Inform. Process. Syst. pp. 29649--29662 (2022)

\bibitem{SSM}
Kundu, J.N., Venkat, N., Babu, R.V., et~al.: Universal source-free domain adaptation. In: IEEE Conf. Comput. Vis. Pattern Recog. pp. 4544--4553 (2020)

\bibitem{DCC}
Li, G., Kang, G., Zhu, Y., Wei, Y., Yang, Y.: Domain consensus clustering for universal domain adaptation. In: IEEE Conf. Comput. Vis. Pattern Recog. pp. 9757--9766 (2021)

\bibitem{SHOT}
Liang, J., Hu, D., Feng, J.: Do we really need to access the source data? source hypothesis transfer for unsupervised domain adaptation. In: Int. Conf. Mach. Learn. pp. 6028--6039 (2020)

\bibitem{UMAD}
Liang, J., Hu, D., Feng, J., He, R.: Umad: Universal model adaptation under domain and category shift  (2021)

\bibitem{BA3US}
Liang, J., Wang, Y., Hu, D., He, R., Feng, J.: A balanced and uncertainty-aware approach for partial domain adaptation. In: Eur. Conf. Comput. Vis. pp. 123--140 (2020)

\bibitem{cross-modal_adaptation}
Lin, Z., Yu, S., Kuang, Z., Pathak, D., Ramanan, D.: Multimodality helps unimodality: Cross-modal few-shot learning with multimodal models. In: IEEE Conf. Comput. Vis. Pattern Recog. pp. 19325--19337 (2023)

\bibitem{STA}
Liu, H., Cao, Z., Long, M., Wang, J., Yang, Q.: Separate to adapt: Open set domain adaptation via progressive separation. In: IEEE Conf. Comput. Vis. Pattern Recog. pp. 2927--2936 (2019)

\bibitem{SPL-OSDA}
Liu, X., Zhou, Y., Zhou, T., Qin, J.: Self-paced learning for open-set domain adaptation. Journal of Computer Research and Development  \textbf{60}(8),  1711--1726 (2023)

\bibitem{AdamW}
Loshchilov, I., Hutter, F.: Decoupled weight decay regularization. In: Int. Conf. Learn. Represent. (2019)

\bibitem{K-means}
MacQueen, J.: Classification and analysis of multivariate observations. In: Berkeley Symp. Math. Statist. Probability. pp. 281--297 (1967)

\bibitem{Pytorch}
Paszke, A., Gross, S., Massa, F., Lerer, A., Bradbury, J., Chanan, G., Killeen, T., Lin, Z., Gimelshein, N., Antiga, L., Desmaison, A., Kopf, A., Yang, E., DeVito, Z., Raison, M., Tejani, A., Chilamkurthy, S., Steiner, B., Fang, L., Bai, J., Chintala, S.: Pytorch: An imperative style, high-performance deep learning library. In: Adv. Neural Inform. Process. Syst. (2019)

\bibitem{DonmaiNet}
Peng, X., Bai, Q., Xia, X., Huang, Z., Saenko, K., Wang, B.: Moment matching for multi-source domain adaptation. In: Int. Conf. Comput. Vis. pp. 1406--1415 (2019)

\bibitem{VisDa}
Peng, X., Usman, B., Kaushik, N., Hoffman, J., Wang, D., Saenko, K.: Visda: The visual domain adaptation challenge  (2017)

\bibitem{GLC}
Qu, S., Zou, T., R\"ohrbein, F., Lu, C., Chen, G., Tao, D., Jiang, C.: Upcycling models under domain and category shift. In: IEEE Conf. Comput. Vis. Pattern Recog. pp. 20019--20028 (2023)

\bibitem{CLIP}
Radford, A., Kim, J.W., Hallacy, C., Ramesh, A., Goh, G., Agarwal, S., Sastry, G., Askell, A., Mishkin, P., Clark, J., et~al.: Learning transferable visual models from natural language supervision. In: Int. Conf. Mach. Learn. pp. 8748--8763 (2021)

\bibitem{Silhouettes}
Rousseeuw, P.J.: Silhouettes: A graphical aid to the interpretation and validation of cluster analysis. Journal of Computational and Applied Mathematics  \textbf{20},  53--65 (1987)

\bibitem{Roy_2019_CVPR}
Roy, S., Siarohin, A., Sangineto, E., Bulo, S.R., Sebe, N., Ricci, E.: Unsupervised domain adaptation using feature-whitening and consensus loss. In: IEEE Conf. Comput. Vis. Pattern Recog. pp. 9471--9480 (2019)

\bibitem{DANCE}
Saito, K., Kim, D., Sclaroff, S., Saenko, K.: Universal domain adaptation through self supervision. In: Adv. Neural Inform. Process. Syst. vol.~33, pp. 16282--16292 (2020)

\bibitem{OVANet}
Saito, K., Saenko, K.: Ovanet: One-vs-all network for universal domain adaptation. In: Int. Conf. Comput. Vis. pp. 9000--9009 (2021)

\bibitem{OSBP}
Saito, K., Yamamoto, S., Ushiku, Y., Harada, T.: Open set domain adaptation by backpropagation. In: Eur. Conf. Comput. Vis. pp. 153--168 (2018)

\bibitem{A_mathematical_theory_of_communication}
Shannon, C.E.: A mathematical theory of communication. The Bell System Technical Journal  \textbf{27}(3),  379--423 (1948)

\bibitem{CoDE}
Shen, M., Lu, Y., Hu, Y., Ma, A.J.: Collaborative learning of diverse experts for source-free universal domain adaptation. In: ACM MM. p. 2054–2065 (2023)

\bibitem{EMA_teacher}
Tarvainen, A., Valpola, H.: Mean teachers are better role models: Weight-averaged consistency targets improve semi-supervised deep learning results. In: Adv. Neural Inform. Process. Syst. (2017)

\bibitem{OfficeHome}
Venkateswara, H., Eusebio, J., Chakraborty, S., Panchanathan, S.: Deep hashing network for unsupervised domain adaptation. In: IEEE Conf. Comput. Vis. Pattern Recog. pp. 5018--5027 (2017)

\bibitem{DDAN}
Wang, J., Chen, Y., Feng, W., Yu, H., Huang, M., Yang, Q.: Transfer learning with dynamic distribution adaptation. {ACM} Trans. Intell. Syst. Technol.  \textbf{11}(1),  6:1--6:25 (2020)

\bibitem{WISE-FT}
Wortsman, M., Ilharco, G., Kim, J.W., Li, M., Kornblith, S., Roelofs, R., Lopes, R.G., Hajishirzi, H., Farhadi, A., Namkoong, H., et~al.: Robust fine-tuning of zero-shot models. In: IEEE Conf. Comput. Vis. Pattern Recog. pp. 7959--7971 (2022)

\bibitem{Zheng2021_IJCV}
Zheng, Z., Yang, Y.: Rectifying pseudo label learning via uncertainty estimation for domain adaptive semantic segmentation. Int. J. Comput. Vision  \textbf{129}(4),  1106–1120 (apr 2021)

\bibitem{CoOp}
Zhou, K., Yang, J., Loy, C.C., Liu, Z.: Learning to prompt for vision-language models. Int. J. Comput. Vis.  \textbf{130}(9),  2337--2348 (jul 2022)

\bibitem{yizhou2023}
Zhou, Y., Bai, S., Zhou, T., Zhang, Y., Fu, H.: Delving into local features for open-set domain adaptation in fundus image analysis. In: MICCAI. pp. 682--692 (2022)

\bibitem{UniAM}
Zhu, D., Li, Y., Yuan, J., Li, Z., Kuang, K., Wu, C.: Universal domain adaptation via compressive attention matching. In: Int. Conf. Comput. Vis. pp. 6974--6985 (2023)

\end{thebibliography}
